\documentclass{article}
\PassOptionsToPackage{numbers}{natbib}

\usepackage[final,main]{neurips_2026}

\usepackage[utf8]{inputenc}
\usepackage[T1]{fontenc}
\usepackage{hyperref}
\usepackage{url}
\usepackage{booktabs}
\usepackage{amsfonts,amsmath,amssymb,amsthm}
\usepackage{thmtools,thm-restate}
\usepackage{nicefrac}
\usepackage{microtype}
\usepackage[table]{xcolor}
\usepackage{multirow}
\usepackage{graphicx}
\usepackage{subcaption}
\usepackage{bm}
\usepackage{bbm}
\usepackage{algorithm}
\usepackage{algorithmic}
\usepackage{enumitem}
\usepackage{colortbl}
\usepackage{array}
\usepackage{mathrsfs}
\setlist[enumerate]{leftmargin=*, itemsep=1pt, topsep=2pt, parsep=0pt}
\newcolumntype{C}[1]{>{\centering\arraybackslash}p{#1}}

\newcommand{\tightbox}[2]{%
  {\setlength{\fboxsep}{0.6pt}\colorbox{#1}{\strut #2}}%
}
\definecolor{hmbase}{HTML}{1F4E8C}

\newcommand{\hmA}[1]{\tightbox{hmbase!80}{\color{white}#1}}
\newcommand{\hmB}[1]{\tightbox{hmbase!58}{\color{white}#1}}
\newcommand{\hmC}[1]{\tightbox{hmbase!36}{#1}}
\newcommand{\hmD}[1]{\tightbox{hmbase!22}{#1}}
\newcommand{\hmE}[1]{\tightbox{hmbase!11}{#1}}
\newcommand{\hmF}[1]{#1}
\theoremstyle{definition}
\newtheorem{definition}{Definition}
\newcommand{\pdn}[2]{\begin{tabular}[c]{@{}c@{}}$#1$\\[-2pt]{\scriptsize $p{=}#2$}\end{tabular}}
\usepackage{mathrsfs}
\newtheorem{proposition}{Proposition}

\newcommand{\ma}{{M_a}}
\newcommand{\mm}{{M_m}}
\newcommand{\lk}{{\mathcal{L}_K}}
\newcommand{\lsep}{{\mathcal{L}_{\text{sep}}}}
\newcommand{\xrgb}{\mathbf{x}}
\newcommand{\xhat}{\hat{\mathbf{x}}}
\newcommand{\xblur}{\tilde{\mathbf{x}}}

\newcommand{\R}{\mathbb{R}}

\newcommand{\sg}{\text{sg}}

\title{Disentangling Where and When:\\
       Factored Spatio-Temporal Explanations for Video Action Recognition}

\author{%
  Bhavan Vasu \\
  Oregon State University \\
  \texttt{vasub@oregonstate.edu}\\
  \And
  Giuseppe Raffa \\
  Intel Labs \\
  \texttt{giuseppe.raffa@intel.com} \\
  \And
    Prasad Tadepalli \\
  Oregon State University \\
  \texttt{prasad.tadepalli@oregonstate.edu} \\
}

\begin{document}

\maketitle
\begin{abstract}
Post-hoc explanation methods for video action recognition typically conflate spatial appearance and temporal dynamics into a single saliency or importance map.  We propose a perturbation-based framework that disentangles these contributions into a spatial appearance mask ($\ma$) and a temporal motion mask ($\mm$). The two masks are jointly optimized against a frozen action recognition model through factored perturbation architecture, with nearest frame replacement to identify important temporal dynamics. A separation penalty discourages the appearance mask from spending budget on frames that the motion mask has already discarded. A ReLU-based area constraint penalizes explanations larger than an assigned budget. To evaluate these explanations, we introduce a separation protocol that sweeps spatial pixels and temporal frames independently, backed by proofs characterizing the proposed method. Across EPIC-Kitchens-55, EGTEA Gaze+, and Something-Something V2, our method attains the lowest deletion AUC against leading baselines while producing masks roughly five times sparser.

\end{abstract}

\section{Introduction}\label{sec:intro}

Video action recognition models such as TSM~\cite{lin2019tsm}, SlowFast~\cite{feichtenhofer2019slowfast}, and Video Swin~\cite{liu2022videoswin} now reach high accuracy on standard benchmarks. They increasingly inform downstream decisions in settings where errors are costly, such as assistive robotics in unstructured home environments~\cite{damen2018scaling}, surgical workflow analysis and skill assessment~\cite{twinanda2016endonet, funke2019video}.  In each of these settings, top-1 accuracy on a held-out test split is a necessary but insufficient guarantee. A practitioner who deploys a video classifier needs to know \emph{why} the model made a given prediction before deciding whether to trust it, deploy it, or audit it for bias.

For static image classification, this question has been studied extensively. Gradient-based methods~\cite{selvaraju2017grad,wang2020score} and perturbation-based methods~\cite{petsiuk2018rise,fong2019extremal} produce spatial saliency maps that highlight image regions correlated with the predicted class. The transition to video introduces a complication that these methods do not address. A video classifier is conditioned on two qualitatively different sources of evidence: the \emph{appearance} of objects within frames, and the \emph{temporal dynamics} across frames. The practical consequences of relying on one source rather than the other are not interchangeable.

\begin{figure}
    \centering
    \includegraphics[height=0.4\linewidth, width=0.8\linewidth]{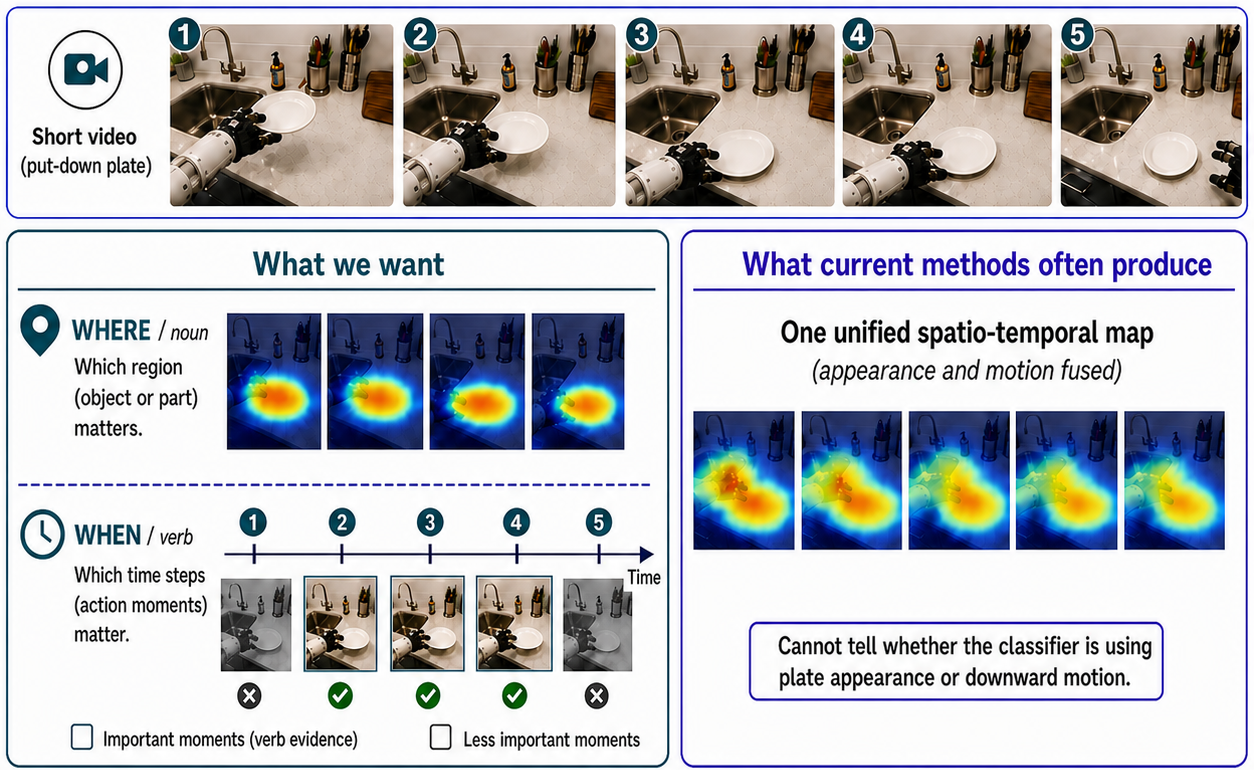}
    \caption{Factored explanations separate appearance (noun: plate)  evidence, where the model looks , from motion evidence (verb: put-down), when the model acts. Standard saliency instead yields a single unified map that conflates appearance and motion.}
    \label{fig:main}
\end{figure}
\paragraph{Motivation:} Consider a kitchen-assistance robot that uses a video classifier to recognize the action ``put-down plate'' from a hand-mounted camera as shown in Figure \ref{fig:main}. There are two qualitatively different ways the classifier might be right. It might recognize the plate (an appearance cue: the noun) and infer the action from context. It might instead recognize the downward trajectory (a motion cue: the verb) and infer the object from context. A standard saliency map cannot distinguish these two cases, but their failure modes differ sharply. A model that keys primarily on plate appearance will misfire when the plate is occluded by the user's hand or replaced by a visually similar distractor (e.g., a flat lid). A model that keys primarily on the downward motion will misfire when the user lowers something that is not a plate, or when the plate is set down in a way that does not match the typical trajectory. The same accuracy figure on a balanced test set hides both failure modes; a factored explanation surfaces them.
\paragraph{Applications:} Disentangling appearance and motion enables critical capabilities across domains. Factored explanations support shortcut auditing by separating spurious appearance correlations (e.g., background bias) from spurious motion correlations (e.g., generic hand sweeps), which single-mask methods conflate. This distinction also supports robustness forecasting: knowing whether a model relies on appearance (vulnerable to occlusion) or motion (vulnerable to viewpoint and framerate) informs deployment constraints and data augmentation. In clinical video analysis, it separates reliance on instrument appearance from procedure-specific motion to help align predictions with expert protocols. In egocentric assistive technology, it distinguishes active user intent (``reaching for $X$'') from passive context (``$X$ is present''). At the dataset level, aggregating masks can reveal static shortcuts, visual ambiguity, and alignment with human attention datasets like EGTEA Gaze+~\cite{li2018egtea} by matching spatial masks to gaze location and temporal masks to fixation timing.

\paragraph{Why existing methods fall short.} Gradient-based methods such as GradCAM~\cite{selvaraju2017grad} and Score-CAM~\cite{wang2020score}, when applied per-frame to a video model, produce spatial heatmaps that conflate spatial and temporal contributions. Aggregating per-frame gradient magnitudes into a temporal importance signal (Temporal GradCAM) preserves time but discards space. Perturbation-based methods adapted to video~\cite{petsiuk2018rise,fong2019extremal,li2021towards} optimize a single spatiotemporal mask, which suffers from the same conflation: the mask is high in regions where \emph{either} the spatial pixel \emph{or} the temporal frame matters, and the user cannot read off which.
Unlike the recent DANCE framework~\cite{Lee2025} that resolves the conflation through an ante-hoc concept bottleneck, our work targets the post-hoc, vocabulary-free regime: factored explanations for any frozen classifier
expressed as 
two coupled but semantically distinct masks. The framework is built around four design choices: 
\begin{enumerate}[leftmargin=*]
\item \textbf{Factored perturbation via Straight-Through Estimator (STE)} (\S\ref{sec:perturbation}). We disentangle evidence into a spatial appearance mask (pixel-level blur) and a temporal motion mask (discrete frame replacement) using 
the STE.  

\item \textbf{One-sided area constraint} (\S\ref{sec:losses}). We replace standard MSE area penalties with a ReLU upper bound, which allows the learned masks to settle at extreme sparsity levels well below the budget.

\item \textbf{Separation loss for optimization degeneracy} (\S\ref{sec:losses}). We introduce a strict penalty that forces the appearance mask to zero on temporally frozen frames. 

\item \textbf{Semantic conditioning and flow alignment} (\S\ref{sec:losses}). We structurally route gradients by supervising the appearance mask with noun confidence and the motion mask with verb confidence. For models lacking dual heads, we substitute semantic conditioning with optical flow alignment.
\end{enumerate}

Standard insertion/deletion metrics~\cite{petsiuk2018rise} evaluate a single mask against a single class probability. Instead, we introduce a \textbf{separation evaluation protocol} (\S\ref{sec:eval}) that sweeps each axis independently: appearance sweeps track \emph{noun} confidence, while motion sweeps track \emph{verb} confidence. We evaluate on EPIC-Kitchens-55~\cite{damen2018scaling}, EGTEA Gaze+~\cite{li2018egtea}, and Something-Something~V2~\cite{goyal2017something} against STEP~\cite{li2021towards}, Extremal Perturbation~\cite{fong2019extremal}, and a unified-mask ablation. Our method achieves the lowest deletion AUC across most datasets and axes, producing masks approximately $5\times$ sparser than the strongest baseline. While Insertion AUC inherently favors the larger masks of our baselines (an artifact of rewarding absolute pixel count, detailed in \S\ref{sec:experiments}), our method attains the highest motion complementary fidelity ($\text{F}^{-}$) across all datasets, directly validating our core disentanglement claim.
\section{Related work: Spatio-temporal explanations for video}\label{sec:related}

Temporal GradCAM aggregates per-frame gradient magnitudes into a one-dimensional importance signal, preserving time at the cost of spatial resolution. SWAG-V~\cite{hartley2022swagv} computes superpixel-weighted gradient maps over video, producing piecewise-constant attributions that are easier to read than dense heatmaps but inherit gradient methods' sensitivity to model nonlinearities. Saliency Tubes~\cite{stergiou2019saliency} reshape the activations of a video network's last 3D convolutional layer into volumetric heatmaps. None of these distinguish spatial from temporal contributions in the resulting map. Perturbation-based methods treat the classifier as a black box and observe confidence change under input modification. STEP~\cite{li2021towards} optimizes a per-frame spatial mask under a Gaussian-blur baseline with an MSE area loss for sparsity. Extremal Perturbation~\cite{fong2019extremal} was originally proposed for images and extends to video by jointly optimizing a smooth fixed-area mask over the spatio-temporal volume.
We use STEP and Extremal Perturbation as perturbation-based baselines in Section~\ref{sec:experiments}; both return a single mask in which spatial and temporal evidence are entangled.

Concept-based methods describe a prediction through higher-level features rather than raw pixels. VTCD~\cite{kowal2024vtcd} discovers reusable spatio-temporal concept tubes via clustering and quantifies their effect on transformer predictions. \citet{ji2023spatiotemporal} extend TCAV to 3D ConvNets, and \citet{saha2024exploring} report that dynamic spatio-temporal concepts carry more information than static ones for action classification. MoTIF~\cite{knab2025concepts} extends concept bottlenecks to video with per-concept temporal self-attention, explaining \emph{when} discovered object and action concepts contribute to a prediction. These approaches test whether a concept influences a prediction without localizing the influence to specific pixels or frames. Language-based variants~\cite{zhuo2019explainable} target semantic-level interpretability through prior knowledge and do not address the appearance motion distinction directly. Contemporaneous to our work, DANCE~\cite{Lee2025} retrains classifiers with a concept bottleneck that separates motion (2D pose clusters) from spatial context (LLM-proposed objects) to attribute predictions to discrete named concepts. Unlike DANCE~\cite{Lee2025}, which retrains models to extract discrete, vocabulary-constrained spatial and motion concepts, our method is \emph{post-hoc} (requires no retraining), outputs \emph{continuous} pixel/frame masks for precise localization, imposes \emph{no vocabulary limits} (handling non-human or occluded actions), and evaluates \emph{empirical} rather than architectural faithfulness.

\section{Method}\label{sec:method}

Let $f: \R^{3 \times T \times H \times W} \to \R^{|\mathcal{V}|} \times \R^{|\mathcal{N}|}$ be a frozen dual-head video classifier producing verb logits $\mathbf{z}^v$ and noun logits $\mathbf{z}^n$ from a clip $\xrgb$ of $T$ uniformly-sampled RGB frames at $H \times W$ resolution. Let $v^* = \arg\max_v \text{softmax}(\mathbf{z}^v(\xrgb))_v$ and $n^* = \arg\max_n \text{softmax}(\mathbf{z}^n(\xrgb))_n$ denote the predicted verb and noun on the clean clip. We seek two masks $\ma \in [0,1]^{T \times H \times W} \text{(appearance: where) and } 
    \mm \in [0,1]^{T} \text{(motion: when)}$, such that:
\begin{align}
    \max_{\ma, \mm} \quad & \text{softmax}(\mathbf{z}^v(\hat{\mathbf{x}}))_{v^*} + \text{softmax}(\mathbf{z}^n(\hat{\mathbf{x}}))_{n^*}
    \label{eq:mainobj}
\end{align}
\vspace{-0.6cm}
\begin{align*}
    \hspace{1cm}\text{s.t.} \quad & \frac{1}{THW} \sum \ma \leq B_a, \quad \frac{1}{T} \sum \mm \leq B_m
\end{align*}

Where $\ma$ identifies the spatial pixels critical for $n^*$ and $\mm$ identifies the temporal frames critical for $v^*$, $B_a, B_m \in (0, 1)$ define the maximum allowable spatial and temporal budgets, forcing the masks to localize minimal subsets of critical evidence. $\ma$ marks the pixels whose preservation (versus a heavy blur) sustains the noun prediction, and $\mm$ marks the frames whose preservation (versus nearest-active-frame replacement) sustains the verb prediction. For single-head models that emit only a single logit vector $\mathbf{z} \in \R^{|\mathcal{C}|}$, the same mask shapes are retained and the noun/verb decomposition is replaced by the optical-flow alignment loss of \S\ref{sec:flow_alignment}. Both masks are parameterized as raw logits and converted to mask values through a temperature-annealed sigmoid $\sigma_\tau(x) = 1/(1 + e^{-\tau x})$, with $\tau$ rising linearly from $\tau_\text{start}{=}1$ to $\tau_\text{max}{=}10$ over the course of optimization. A low temperature in early iterations preserves gradient flow; a high temperature in later iterations sharpens the mask toward $\{0,1\}$. The appearance mask is optimized at low resolution $H_s = W_s = \lfloor H/r \rfloor$ with $r = 7$, then bilinearly upsampled to $H \times W$, Gaussian-smoothed (kernel size $2r-1$, $\sigma = r/2$), and clamped to $[0,1]$. The motion mask is parameterized as $T$ scalar logits, one per frame.
\subsection{Factored perturbation with hard frame selection}\label{sec:perturbation}

The perturbation is a single compositional forward pass with two factors: a spatial blur based blend governed by the appearance mask $\ma$, and a temporal nearest-frame replacement governed by the motion mask $\mm$ that consumes the spatially-perturbed clip as its replacement source. Both masks are parameterized by logits $(\boldsymbol{\ell}_A,\boldsymbol{\ell}_M)$ and optimized \emph{jointly} under a single objective (Sec.~\ref{sec:losses}); the ``factor'' ordering below describes the forward composition, not a staged optimization schedule.

\paragraph{Spatial factor: blur--blend.} Given the clean clip $\xrgb$ and a heavily blurred copy $\xblur = \mathcal{G}_B * \xrgb$ (kernel size $k_B{=}31$, $\sigma_B{=}20$),
\begin{equation}
    \mathbf{x}^{\text{app}} = \ma \odot \xrgb + (1 - \ma) \odot \xblur.
\end{equation}
Blur is preferred to a zero or mean-frame baseline because it preserves low-frequency context, keeping the perturbed clip closer to the model's training distribution and reducing the risk that observed confidence drops are attributable to distribution shift rather than removed evidence.

\paragraph{Temporal factor: nearest-frame replacement.} Frames that the motion mask designates as inactive are replaced by the content of the nearest \emph{active} frame from $\mathbf{x}^{\text{app}}$. The replacement source is the spatially-perturbed clip rather than the original, so the appearance mask of the source frame is propagated to its replacement targets; the choice of replacement source is discrete in the forward pass. 

Concretely, let $\mm = (m_1, \ldots, m_T)$. The proximity-weighted, motion-gated replacement weights
\begin{equation}
    w_{i \to j} = \frac{\exp(-\|i - j\|^2 / 2\sigma_R^2) \cdot m_j}{\sum_{k \neq i} \exp(-\|i - k\|^2 / 2\sigma_R^2) \cdot m_k + \epsilon}, \qquad \sigma_R = 2.
\end{equation}
A pure soft average over $\{w_{i\to j}\}$ produces a blurred replacement frame, reintroducing the artifact the temporal factor was intended to avoid. We instead take the argmax in the forward pass and let gradients flow through the soft weights via a Straight-Through Estimator:
\begin{equation}
    j^*(i) = \arg\max_j w_{i \to j}, \qquad
    \tilde{w}_{i \to j} = \mathbbm{1}[j = j^*(i)] - \sg(w_{i \to j}) + w_{i \to j}.
\end{equation}
A second STE binarizes the temporal gate itself,
\begin{equation}
    \mm^{\textnormal{hard}}_t = \mathbbm{1}[m_t > 0.5], \qquad
    \mm^{\textnormal{STE}}_t = \mm^{\textnormal{hard}}_t - \sg(m_t) + m_t,
\end{equation}
and the final perturbed clip is
\begin{equation}\label{eq:full_perturbation}
    \xhat_t = \mm^{\text{STE}}_t \cdot \mathbf{x}^{\text{app}}_t + (1 - \mm^{\text{STE}}_t) \cdot \sum_j \tilde{w}_{t \to j} \, \mathbf{x}^{\text{app}}_j.
\end{equation}
At convergence, frozen frames duplicate the nearest active frame to structurally remove temporal dynamics. Preserving gradient flow via a Straight-Through Estimator (STE) introduces a visual discrepancy between the true discrete selection ($\xhat^{\textnormal{hard}}_i$) and its continuous proxy ($\xhat^{\textnormal{soft}}_i$). 

\begin{restatable}[STE bias bound]{proposition}{propSTE}\label{prop:ste_bias}
If $\Delta_i$ is the probability gap between the top two replacement candidates, and let $\delta_i$ is the maximum visual discrepancy among candidates. Then, the pointwise discrepancy between $\xhat^{\textnormal{hard}}_i$ and $\xhat^{\textnormal{soft}}_i$ is bounded by (proof in App.~\ref{app:thm:ste})
:
\[
\|\xhat^{\textnormal{hard}}_i - \xhat^{\textnormal{soft}}_i\| \;\leq\; (1 - \Delta_i)\delta_i
\]
\end{restatable}

Empirically, temperature annealing drives the mask toward a discrete one-hot state ($\Delta_i \to 1$), safely collapsing this discrepancy to zero. Choosing $\mathbf{x}^{\text{app}}$ as the replacement tightens the STE bias bound of Proposition~\ref{prop:ste_bias}: the visual discrepancy $\delta_i$ between candidate replacement frames is smaller for spatially-cleaned frames than for raw frames, since $\ma$ removes irrelevant background variation, so the bound $(1-\Delta_i)\delta_i$ contracts.

\par Alternative strategies for replacing temporally inactive frames introduce severe artifacts: zero or mean baselines cause distribution shift, frame shuffling creates spurious trajectories (failing on circular motions), and soft blending leaves residual ghosting. By replacing inactive frames with exact copies of their nearest active neighbors, the method creates static blocks of identical frames. This completely halts the motion trajectory, reducing temporal dynamics to exactly zero rather than merely attenuating them, while preserving spatial appearance. Crucially, this nearest-neighbor approach minimizes the visual discrepancy factor $\delta_i$, directly tightening the formal STE bias bound (Proposition~\ref{prop:ste_bias}).

\subsection{Loss functions}\label{sec:losses}

For dual-head models, the total loss is
\begin{equation}\label{eq:total_loss}
    \mathcal{L} = \underbrace{-\log p_{n^*}(\xhat) - \log p_{v^*}(\xhat)}_{\text{split prediction}}
        + \underbrace{\lambda_A \mathcal{L}_{\text{area}}^a + \lambda_M \mathcal{L}_{\text{area}}^m}_{\text{sparsity}}
        + \underbrace{\lambda_S \lk}_{\text{smoothness}}
        + \underbrace{\lambda_D \lsep}_{\text{separation}}.
\end{equation}

\paragraph{Split prediction.} The noun log-probability $-\log(\text{softmax}(\mathbf{z}^n(\xhat))_{n^*} )$ supervises the appearance mask, and the verb log-probability supervises the motion mask. The logarithmic form is preferred to the linear form $-p$ because the gradient of $-\log p$ does not vanish as $p \to 0^+$, which accelerates convergence in early iterations when the masks are nearly empty and target-class probability is consequently low. Single-head models add the flow alignment loss of \S\ref{sec:flow_alignment}, and replace the split prediction with a single target-class log-probability.

Measured at convergence, cross-axis gradients are of the same order as the intended ones (median leak ratio $0.85$ for $\ma$ and $1.23$ for $\mm$; App.~\ref{app:leakage}). Semantic separation of the explanations is therefore an empirical outcome that depends on the classifier's learned representations and on the available supervision; removing the noun/verb split degrades deletion AUC and $\text{F}^{-}$ on both axes (App.~\ref{app:semantic_split}).

\paragraph{Sparsity constraint.} Prior work~\cite{li2021towards} realizes the sparsity constraint in Equation \ref{eq:mainobj} through an MSE loss that penalizes the sorted mask values for deviating from a binary target with $B_a$ or $B_m$ active entries. This penalty is symmetric: the mask is penalized for using \emph{fewer} pixels than the budget as well as more, which forces the optimizer to occupy the budget exactly regardless of input complexity. We argue that an explanation should be as sparse as possible within the budget. Replacing the MSE with a ReLU upper bound,
\begin{equation}\label{eq:relu_area}
    \mathcal{L}_{\text{area}}^{\text{ReLU}}(\mathbf{m}, v) = \frac{1}{N}\sum_{i=1}^{N} \big[\text{ReLU}(\text{sort}(\mathbf{m})_i - t_i)\big]^2,
\end{equation}
penalizes only excess. Empirically, the realized appearance mask settles at roughly $5\times$ smaller than the budget on EPIC-Kitchens-55 and EGTEA (Table~\ref{tab:dataset_model_results}), whereas under the symmetric MSE form the mask occupies the budget exactly. Appendix~\ref{app:thm:area} formalizes this mathematical distinction: under standard weak monotonicity assumptions on the prediction loss landscape, the symmetric MSE form provably forces the realized area to saturate the budget at a rate of $\mathcal{O}(1/\lambda)$ (Proposition~\ref{prop:mse_saturation}). 
\begin{proposition}[MSE forces budget saturation, with quantitative rate]\label{prop:mse_saturation}
Let $\mathbf{m}^*_\lambda$ be an interior stationary point of $\mathcal{L}_{\textnormal{pred}} + \lambda \mathcal{L}_{\textnormal{area}}^{\textnormal{MSE}}$. As $\lambda \to \infty$, the realized area $a^*_\lambda$ converges to the target budget $b$, and the gap closes at a rate of $O(1/\lambda)$ with a constant scaling of $\sqrt{N}$. Specifically, the realized area satisfies the following bound:
\[
|a^*_\lambda - b| \;\leq\; \frac{L \sqrt{N}}{2\lambda}
\]
\end{proposition}
Conversely, the ReLU formulation analytically permits sub-budget stationary points, allowing the realized sparsity to be dictated entirely by the underlying prediction-loss geometry independent of the penalty weight $\lambda$ (proof App.~\ref{app:thm:area}).

\begin{proposition}[ReLU permits sub-budget stationary points independent of $\lambda$]
Suppose there exists a mask $\mathbf{m}^* \in [0,1]^N$ that is a stationary point of the prediction loss ($\nabla \mathcal{L}_{\textnormal{pred}}(\mathbf{m}^*) = \mathbf{0}$) and naturally satisfies the upper-bound budget constraint ($\mathbf{m}^{\textnormal{sort},*}_i \leq t_i$ for all $i$, yielding a realized area $a^* \leq b$). Then, $\mathbf{m}^*$ remains a valid stationary point of the joint loss for every penalty weight $\lambda > 0$, satisfying:
\[
\nabla \Big( \mathcal{L}_{\textnormal{pred}}(\mathbf{m}^*) + \lambda \mathcal{L}_{\textnormal{area}}^{\textnormal{ReLU}}(\mathbf{m}^*, v) \Big) = \mathbf{0}
\]
\end{proposition}

\paragraph{Ellipsoid-kernel smoothness.} Following \cite{li2021towards}, we convolve $\ma$ with a 3D ellipsoid kernel ($t_k{=}3$, $h_k{=}w_k{=}31$, supported on the unit ball) and apply the same sorted-target loss as the sparsity term. This regularizer encourages the mask to form connected spatiotemporal regions rather than isolated high-activation pixels.

\paragraph{Separation.} The separation loss penalizes appearance activation in temporally frozen frames:
\begin{equation}\label{eq:disent}
    \lsep = \frac{1}{T H W}\sum_{t,h,w} M_{a,t,h,w} \cdot (1 - M_{m,t}).
\end{equation}
The motivation is precise: when $M_{m,t} \approx 0$, frame $t$ is replaced by content from a nearby active frame, so any spatial highlights $\ma$ assigns to the original frame $t$ are never observed by the model. Without an explicit penalty, the optimizer is free to expend appearance budget on these unobserved positions. With Eq.~\eqref{eq:disent}, the appearance mask concentrates on the frames that survive temporal gating. 

\subsection{Flow alignment for single-head models}\label{sec:flow_alignment}

Something-Something~V2 contains 174 composite action classes that admit no natural verb/noun decomposition (e.g., \emph{pushing something from left to right} versus \emph{pushing something from right to left}). The split prediction loss therefore has nothing to split. To drive $\ma$ and $\mm$ toward semantically distinct quantities, we anchor the $M_m$ to a physics-grounded signal: per-frame optical flow magnitude.

Let $\bar{\phi}_t = \|\text{FlowMag}(t)\| / \max_t \|\text{FlowMag}(t)\| \in [0,1]$ denote the normalized flow magnitude at frame $t$, computed via \cite{farneback2003two}. The flow alignment loss is
\begin{equation}
    \mathcal{L}_{\text{flow}} = 1 - \cos(\mm,\, \bar{\phi}).
\end{equation}
This biases $\mm$ toward frames in which pixels are observably moving, while leaving the prediction loss (a single $-\log p_{c^*}$) free to drive the appearance mask. We acknowledge this represents a pragmatic empirical heuristic rather than a structurally guaranteed decoupling; however, anchoring the temporal axis to explicit optical flow provides the necessary inductive bias to separate temporal dynamics from static spatial features when independent semantic verb/noun logits are unavailable. Flow alignment is a real inductive bias rather than a neutral regularizer. The learned $\mm$ is only moderately correlated with flow magnitude ($r = 0.37$), so it is influenced by, but does not reproduce, the flow signal. Our strongest appearance/motion separation results are therefore on the dual-head datasets, where the separation is driven by the model's own noun and verb outputs.

\section{Evaluation protocol}\label{sec:eval}

Standard insertion/deletion metrics~\cite{petsiuk2018rise} sweep a single mask in order of importance and track one class probability. For a factored explanation, this answers the wrong question: it reports how much confidence is recovered from the union of spatial and temporal cues, not how much each axis contributes. We propose a separation protocol that treats the two axes independently.

\paragraph{Appearance axis.} Sort spatial positions in descending order of $\ma$, with frames outside $\mm$ kept frozen. For a fraction $\alpha$, keep the top-$\alpha$ positions sharp and blur the rest, and track the noun (dual-head) or class (single-head) confidence:
\begin{equation}
    \text{Ins}_a(\alpha) = \text{softmax}\!\left(\mathbf{z}^n\!\left(\ma^{(\alpha)} \odot \xrgb + (1-\ma^{(\alpha)}) \odot \xblur\right)\right)_{n^*}.
\end{equation}
Deletion is defined symmetrically using the complement mask. The area under the insertion curve $\text{AUC}^a_{\text{Ins}} = \int_0^1 \text{Ins}_a(\alpha)\, d\alpha$ is to be maximized; the area under the deletion curve is to be minimized.

\paragraph{Motion axis.} Sort frames in descending order of $\mm$. Starting from the $\ma$-masked clip, un-freeze (insertion) or freeze (deletion) frames in that order using the nearest-active-frame replacement, and track the verb or class confidence. The two AUCs $\text{AUC}^m_{\text{Ins}}$ and $\text{AUC}^m_{\text{Del}}$ are interpreted as on the appearance axis.

\paragraph{Fidelity.} $\text{F}^{+}$ is the fraction of clips for which the masked input preserves the model's original top-1 prediction; $\text{F}^{-}$ is the fraction for which the \emph{complement} of the masked input changes it. Both are to be maximized, per axis. $\text{F}^{+}$ asks whether the mask retains the evidence; $\text{F}^{-}$ asks whether the mask retains \emph{only} the evidence. The latter is the more direct test of explanation faithfulness, since recovering confidence with a large mask is easier than destroying it with a small one.

\paragraph{Sparsity.} $\overline{A}_a$ and $\overline{A}_m$ denote the mean active fractions of pixels and frames, averaged over the test set. Because Eq.~\eqref{eq:relu_area} permits below-budget operation, $\overline{A}$ may fall well below the optimization target. Standard deviations are reported as well: high variance is itself diagnostic, indicating that some action classes require substantially more evidence than others.

\begin{table*}[t]
\centering
\caption{Explanation quality comparison. Each cell reports TSM/TimeSformer. App/Mot denote appearance/motion; Ins/Del denote insertion/deletion AUC. $F^+$ and $F^-$ are fidelity metrics; $M_a$, $M_m$ are active pixel/frame fractions. Color intensity within each dataset/model column maps to relative performance (darker = better, computed by min--max normalization, bucketed into 6 levels). Paired significance tests and standard deviations are reported in Appendix~\ref{app:stats}.}
\label{tab:dataset_model_results}
\small
\setlength{\tabcolsep}{2.0pt}
\renewcommand{\arraystretch}{1.2}
\begin{tabular}{@{}l*{8}{C{0.072\textwidth}}*{2}{C{0.080\textwidth}}@{}}
\toprule
& \multicolumn{2}{c}{\textbf{Insertion AUC}}
& \multicolumn{2}{c}{\textbf{Deletion AUC}}
& \multicolumn{2}{c}{\textbf{Fidelity+}}
& \multicolumn{2}{c}{\textbf{Fidelity$-$}}
& \multicolumn{2}{c}{\textbf{Mask size}} \\
\cmidrule(lr){2-3} \cmidrule(lr){4-5} \cmidrule(lr){6-7}
\cmidrule(lr){8-9} \cmidrule(lr){10-11}
\textbf{Method}
& App$\uparrow$ & Mot$\uparrow$
& App$\downarrow$ & Mot$\downarrow$
& App$\uparrow$ & Mot$\uparrow$
& App$\uparrow$ & Mot$\uparrow$
& $M_a\downarrow$ & $M_m\downarrow$ \\
\midrule
\multicolumn{11}{c}{\textbf{EGTEA (N = 400), TSM/TimeSformer}} \\
\midrule
EP + $M_m$
& \hmE{.72}/\hmF{.65} & \hmF{.79}/\hmF{.71}
& \hmD{.41}/\hmB{.24} & \hmF{.77}/\hmE{.67}
& \hmF{.87}/\hmF{.85} & \hmF{.87}/\hmF{.80}
& \hmF{.14}/\hmD{.40} & \hmF{.11}/\hmF{.17}
& \hmE{50.5}/\hmE{50.7} & \hmD{61.6}/\hmD{61.6} \\
EP
& \hmF{.70}/\hmF{.65} & --
& \hmD{.43}/\hmB{.27} & --
& \hmE{.91}/\hmB{.96} & --
& \hmF{.12}/\hmE{.33} & --
& \hmE{50.6}/\hmE{50.8} & -- \\
STEP + $M_m$
& \hmC{.77}/\hmC{.72} & \hmA{.84}/\hmC{.82}
& \hmE{.46}/\hmE{.37} & \hmF{.79}/\hmF{.74}
& \hmB{.97}/\hmB{.97} & \hmB{.96}/\hmA{.96}
& \hmA{.43}/\hmA{.72} & \hmF{.12}/\hmF{.20}
& \hmF{60.6}/\hmF{69.0} & \hmF{81.8}/\hmF{83.3} \\
STEP
& \hmF{.70}/\hmB{.73} & --
& \hmF{.51}/\hmF{.41} & --
& \hmA{.99}/\hmA{.99} & --
& \hmC{.28}/\hmB{.57} & --
& \hmF{56.4}/\hmF{65.4} & -- \\
\midrule
\textbf{Unified Mask}
& \hmA{.81}/\hmA{.78} & --
& \hmD{.43}/\hmE{.36} & --
& \hmA{1.00}/\hmA{1.00} & --
& \hmF{.15}/\hmF{.23} & --
& \hmA{20.2}/\hmA{18.2} & -- \\
\textbf{Ours (Full)}
& \hmD{.75}/\hmF{.63} & \hmF{.79}/\hmA{.88}
& \hmA{.27}/\hmA{.20} & \hmA{.25}/\hmA{.35}
& \hmA{.99}/\hmA{.99} & \hmA{.99}/\hmA{.99}
& \hmD{.25}/\hmD{.41} & \hmA{.82}/\hmA{.65}
& \hmA{12.4}/\hmA{10.7} & \hmA{37.9}/\hmA{33.6} \\
\midrule
\multicolumn{11}{c}{\textbf{EPIC-Kitchen-55 (N = 9385), TSM/TimeSformer}} \\
\midrule
EP + $M_m$
& \hmF{.40}/\hmA{.54} & \hmF{.53}/\hmF{.39}
& \hmC{.15}/\hmC{.12} & \hmE{.50}/\hmC{.37}
& \hmF{.54}/\hmF{.41} & \hmF{.33}/\hmF{.43}
& \hmD{.57}/\hmE{.23} & \hmF{.20}/\hmC{.08}
& \hmD{50}/\hmD{50.3} & \hmD{61.5}/\hmD{61.4} \\
EP
& \hmF{.41}/\hmF{.31} & --
& \hmD{.18}/\hmD{.14} & --
& \hmF{.58}/\hmE{.54} & --
& \hmF{.30}/\hmE{.22} & --
& \hmE{50.4}/\hmD{49.2} & -- \\
STEP + $M_m$
& \hmA{.58}/\hmC{.45} & \hmA{.70}/\hmC{.61}
& \hmE{.20}/\hmF{.18} & \hmF{.62}/\hmF{.54}
& \hmB{.89}/\hmB{.80} & \hmD{.65}/\hmC{.75}
& \hmA{.87}/\hmA{.42} & \hmF{.21}/\hmF{.02}
& \hmF{70.2}/\hmF{75.2} & \hmF{89.8}/\hmF{86.8} \\
STEP
& \hmA{.57}/\hmC{.43} & --
& \hmF{.23}/\hmF{.17} & --
& \hmA{.94}/\hmB{.88} & --
& \hmD{.55}/\hmB{.37} & --
& \hmF{66.1}/\hmF{73.3} & -- \\
\midrule
\textbf{Unified Mask}
& \hmA{.59}/\hmD{.39} & --
& \hmF{.22}/\hmE{.16} & --
& \hmA{.99}/\hmA{.99} & --
& \hmF{.28}/\hmF{.18} & --
& \hmB{21.1}/\hmA{19.6} & -- \\
\textbf{Ours (Full)}
& \hmE{.44}/\hmF{.29} & \hmB{.66}/\hmA{.77}
& \hmA{.10}/\hmA{.08} & \hmA{.21}/\hmA{.25}
& \hmA{.97}/\hmA{.96} & \hmA{.98}/\hmA{.96}
& \hmE{.44}/\hmD{.26} & \hmA{.62}/\hmA{.29}
& \hmA{10.3}/\hmA{9.6} & \hmA{24.5}/\hmA{27.6} \\
\midrule
\multicolumn{11}{c}{\textbf{SSv2 (N = 200), TSM/TimeSformer}} \\
\midrule
EP + $M_m$
& \hmF{.26}/\hmE{.42} & \hmC{.40}/\hmF{.47}
& \hmA{.08}/\hmA{.13} & \hmF{.39}/\hmD{.47}
& \hmF{.53}/\hmF{.62} & \hmC{.55}/\hmD{.62}
& \hmE{.40}/\hmE{.39} & \hmF{.47}/\hmB{.46}
& \hmD{52.0}/\hmD{51.9} & \hmD{64.4}/\hmD{64.4} \\
EP
& \hmD{.32}/\hmC{.48} & --
& \hmD{.14}/\hmC{.18} & --
& \hmC{.83}/\hmC{.84} & --
& \hmF{.28}/\hmF{.29} & --
& \hmD{52.0}/\hmD{51.9} & -- \\
STEP + $M_m$
& \hmB{.36}/\hmA{.56} & \hmA{.48}/\hmA{.58}
& \hmE{.15}/\hmD{.22} & \hmF{.44}/\hmF{.56}
& \hmA{.93}/\hmA{.95} & \hmA{.93}/\hmA{.94}
& \hmA{.66}/\hmA{.68} & \hmA{.61}/\hmA{.48}
& \hmF{79.3}/\hmF{78.8} & \hmF{86.7}/\hmF{85.5} \\
STEP
& \hmC{.34}/\hmB{.52} & --
& \hmF{.17}/\hmE{.24} & --
& \hmA{.99}/\hmA{.99} & --
& \hmA{.64}/\hmB{.60} & --
& \hmF{78.9}/\hmF{78.7} & -- \\
\midrule
\textbf{Unified Mask}
& \hmA{.39}/\hmC{.47} & --
& \hmF{.18}/\hmF{.29} & --
& \hmA{1.00}/\hmA{1.00} & --
& \hmA{.63}/\hmC{.54} & --
& \hmD{51.1}/\hmD{51.6} & -- \\
\textbf{Ours (Full)}
& \hmD{.31}/\hmF{.37} & \hmF{.25}/\hmF{.46}
& \hmB{.11}/\hmA{.10} & \hmA{.11}/\hmA{.32}
& \hmA{.93}/\hmB{.90} & \hmF{.13}/\hmF{.41}
& \hmA{.60}/\hmC{.52} & \hmA{.62}/\hmF{.42}
& \hmA{22.8}/\hmA{24.0} & \hmA{22.1}/\hmA{26.2} \\
\bottomrule
\end{tabular}
\end{table*}
\section{Experiments}\label{sec:experiments}

\subsection{Setup}

\paragraph{Datasets and Models} We evaluate on three benchmarks. EPIC-Kitchens-55~\cite{damen2018scaling} provides egocentric kitchen video with $(verb, noun)$ pair labels and serves as our primary dual-head testbed. EGTEA Gaze+~\cite{li2018egtea} provides a smaller egocentric corpus with synchronized gaze recordings with $(verb, noun)$ pair labels. Something-Something~V2~\cite{goyal2017something} contains 174 single-label composite classes whose discrimination requires temporal reasoning, and is the testbed for the flow-alignment variant. Table \ref{tab:dataset_model_results} presents results on a randomly picked subset (listed in table) of the three datasets. For EPIC-Kitchens-55 and EGTEA Gaze+ we use frozen TSM~\cite{lin2019tsm} (ResNet-50) and TimeSformer~\cite{gberta_2021_ICML} models with separate verb and noun heads and for Something-Something~V2 we use the single-head versions of the same backbones.

\subsection{Qualitative results}
\label{ap:qualitative}

\begin{figure}[h!]
    \centering
    \includegraphics[width=\linewidth]{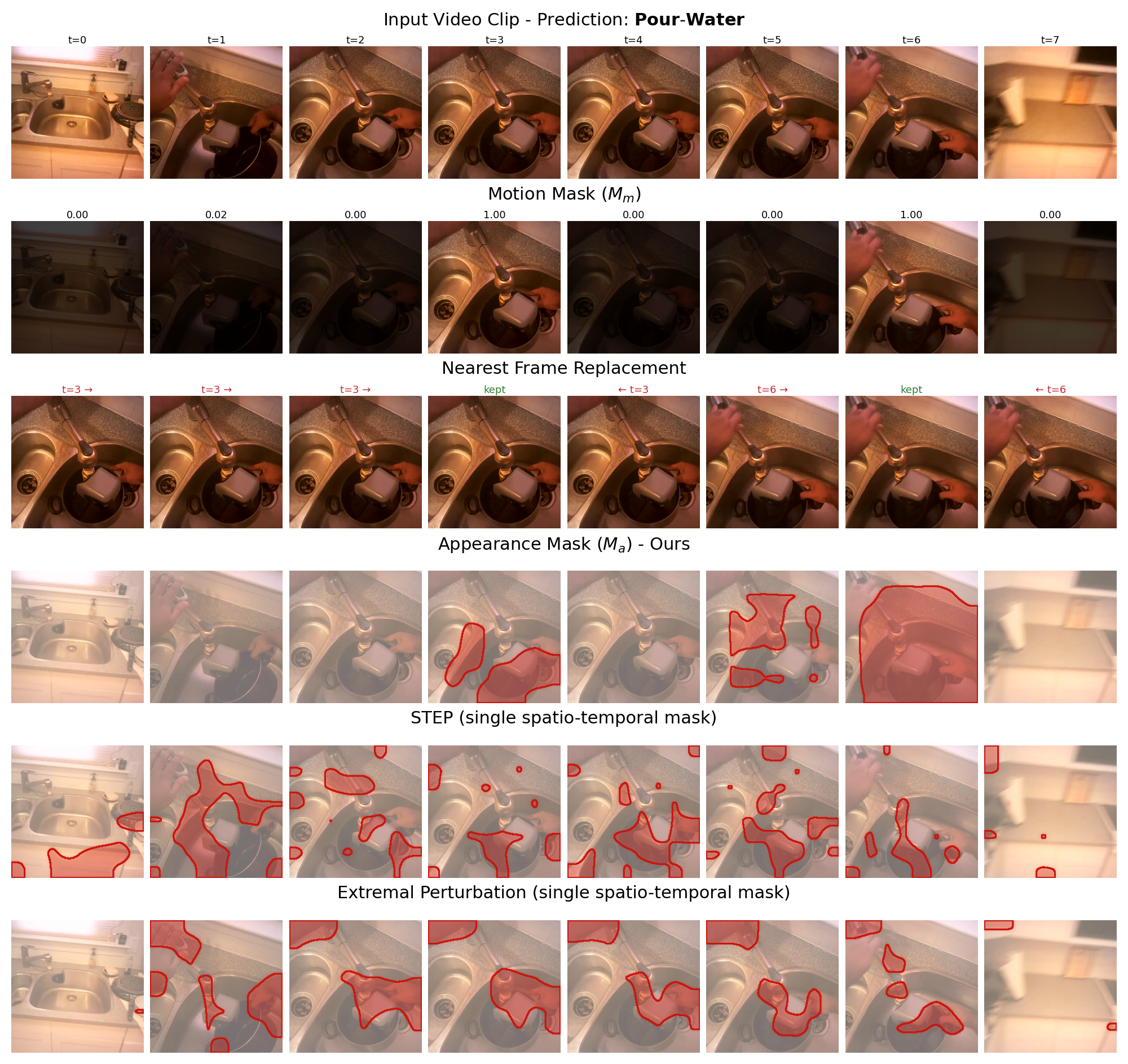}
    
    \caption{Qualitative figure showing (a) Input video clip; (b) Motion mask $M_m$; (c) The video clip with the winning nearest-frame replacement that resulted in $M_m$; (d) Appearance mask $M_a$ produced by our approach, STEP and EP. Highlighted regions in $M_a$ show the top 15\% of important regions.}
    \label{fig:qual1}
\end{figure}

\paragraph{Baselines.} Extremal Perturbation~\cite{fong2019extremal}, STEP~\cite{li2021towards} and a Unified Mask ablation that optimizes a single $\ma' \in [0,1]^{T\times H\times W}$ within our perturbation pipeline but without the appearance/motion factorization. The unified ablation isolates the contribution of the factored design itself, controlling for the perturbation baseline, optimization schedule, and area-loss formulation.

\paragraph{Implementation details.} Optimization runs for 500 steps of Adam with learning rate 0.05 on $T = 8$ uniformly sampled frames at $224 \times 224$ resolution. The loss weights $(\lambda_A, \lambda_M, \lambda_S, \lambda_D)$ are set via a coarse grid search and then a local Optuna search around the best region on a held-out validation subset. We compare methods with per-clip paired tests (Wilcoxon signed-rank). Hyperparameters, sensitivity sweeps, hardware, and timing are reported in the Appendix.

\subsection{Results}
Table~\ref{tab:dataset_model_results} reports the separation metrics across the three benchmarks. Figure~\ref{fig:qual1} shows an EGTEA Gaze+ clip that a TSM model classifies as \emph{pour water}. The motion mask keeps two of the eight frames ($t{=}3$ and $t{=}6$), in which the hand operates the faucet above the pot, and discards the frames where the camera shows the whole sink ($t{=}0$) or turns away ($t{=}7$); each discarded frame is replaced by a copy of its nearest kept frame. The appearance mask appears only on the three frames out of which two frames are kept by $M_m$ and cover the faucet, the pot and the hand. STEP and Extremal Perturbation, which use a single spatio-temporal mask, place their evidence on different regions in nearly every frame, including the frames showing the whole sink and the camera turning away.

\paragraph{Deletion AUC.} With the TSM backbone, our method attains the lowest motion deletion AUC on all three datasets and the lowest appearance deletion AUC on EPIC-Kitchens-55 and EGTEA. On EPIC-Kitchens-55, appearance deletion AUC drops to $0.10$ versus $0.18$ for Extremal Perturbation and $0.23$ for STEP, and motion deletion AUC drops to $0.21$ versus $0.50$ for the Extremal Perturbation temporal extension (EP + $M_m$). On EGTEA the values are $0.27$ (appearance) and $0.25$ (motion). On Something-Something~V2, motion deletion AUC is again the lowest ($0.11$ versus $0.39$ for EP + $M_m$), but appearance deletion AUC ($0.11$) is second to EP + $M_m$ ($0.08$) and statistically tied with Extremal Perturbation (Table~\ref{tab:significance}). With the TimeSformer backbone, our method has the lowest deletion AUC on every dataset along both axes. Removing the regions and frames selected by our masks therefore lowers target-class confidence more than removing those selected by the baselines in all but one setting, indicating that the masks capture evidence the classifier relies on.

\paragraph{Mask sparsity.} On EPIC-Kitchens-55, our appearance mask retains only $10.3\%$ of pixels and our motion mask retains only $24.5\%$ of frames. STEP retains $66.1\%$ of pixels and lacks a temporal mask (using every frame); Unified Mask retains $21.1\%$ of pixels but similarly uses every frame. The pattern recurs on EGTEA ($12.4\%$ appearance / $37.9\%$ motion for our method versus $56.4\%$ and all frames for STEP) and on Something-Something~V2 ($22.8\%$ / $22.1\%$ versus $78.9\%$ and all frames). The ReLU upper-bound area loss therefore performs as designed: when the input does not require its full budget, the optimizer uses less.

\paragraph{Insertion AUC.} Our method does not lead on insertion AUC. Baselines with larger masks, such as the STEP variants and Unified Mask, often attain higher insertion scores. On EGTEA, Unified Mask achieves an appearance insertion AUC of $0.81$ and STEP + $M_m$ reaches $0.84$ on the motion axis, compared to our $0.75$ and $0.79$. On EPIC-Kitchens-55, STEP + $M_m$ reaches $0.70$ motion insertion versus our $0.66$. On Something-Something~V2, Unified Mask leads appearance insertion ($0.39$) and STEP + $M_m$ leads motion insertion ($0.48$), against our $0.31$ and $0.25$. As formalized in Proposition~\ref{prop:ins_advantage}, for binary or near-binary masks insertion AUC tends to reward mask size: a larger mask that contains the evidence keeps confidence high over a longer portion of the sweep. Our appearance masks retain $3.5$--$6.4\times$ fewer pixels than STEP and about $2\times$ fewer than Unified Mask, so they are at a disadvantage under this metric, which is why we treat deletion AUC as the primary measure of evidence localization.

\begin{proposition}[Insertion AUC favors larger faithful masks when they are forced to be binary i.e important or not]\label{prop:ins_advantage}
Assuming two masks $M^{(1)}$ and $M^{(2)}$, characterized by active area fractions $0 < a_1 < a_2 < 1$ and strictly satisfy the faithfulness axioms of Definition~\ref{def:faithful}, and assuming the slope of the insertion curve of $M^{(2)}$ over the interval $[a_1, a_2]$ is bounded below by a strictly positive constant $\sigma > 0$. Then the difference between insertion AUC is given by:
\begin{equation}\label{eq:ins_lower}
\textnormal{AUC}_{\textnormal{Ins}}^{(2)} - \textnormal{AUC}_{\textnormal{Ins}}^{(1)}
\;\geq\; \tfrac{1}{2} \sigma \, (a_2 - a_1)^2.
\end{equation}
The accumulated gap remains strictly positive under any circumstance where method $2$'s supplementary positions contain individually informative predictive signals.
\end{proposition}

\begin{proposition}[Deletion AUC depends on coverage, not size]\label{prop:del_invariance} Suppose $M^{(1)}$ sufficiently covers the model's underlying evidence, where the deletion curve $D^{(1)}$ perfectly converges to the baseline confidence: $D^{(1)}(a^*) = C_{\textnormal{base}}$(fully blurred video). Under these conditions:
$\textnormal{AUC}_{\textnormal{Del}}^{(2)} \;=\; \textnormal{AUC}_{\textnormal{Del}}^{(1)}.$
\end{proposition}
Appendix~\ref{app:thm:ins_del} formalizes this argument: under perfect-faithfulness assumptions, a larger mask provably enjoys higher insertion AUC by a margin linear in the area gap (Proposition~\ref{prop:ins_advantage}), while deletion AUC is invariant to mask size at fixed precision (Proposition~\ref{prop:del_invariance}). Our method outperforms or matches all baselines on deletion AUC across all three datasets and along both axes, which we view as the more direct evidence of explanation quality supported by the above propositions. 

\paragraph{Explanation fidelity ($\text{F}^{+}$ and $\text{F}^{-}$).} We evaluate faithfulness via evidence retention ($\text{F}^{+}$) and complementary fidelity ($\text{F}^{-}$). With the TSM backbone, our method attains the highest motion $\text{F}^{-}$ on all three datasets: $0.82$ on EGTEA (versus $0.11$ for EP + $M_m$ and $0.12$ for STEP + $M_m$), $0.62$ on EPIC-Kitchens-55 (versus $0.20$ and $0.21$), and $0.62$ on Something-Something~V2, where STEP + $M_m$ is close ($0.61$). Removing the frames selected by $M_m$ therefore lowers confidence in the target (verb or class) far more than removing the frames selected by the baselines' temporal masks, indicating that $M_m$ captures the temporal evidence the classifier relies on. Although our appearance masks retain $3.5$--$6.4\times$ fewer pixels than STEP, they keep $\text{F}^{+}\geq 0.97$ on EPIC-Kitchens-55 and EGTEA. The exception is motion $\text{F}^{+}$ on Something-Something~V2 ($0.13$): keeping only the ${\sim}22\%$ of frames selected by $M_m$ is not sufficient to preserve the prediction there, consistent with the lower motion insertion AUC on that dataset.

\paragraph{Ablations.}
Isolating our factorization, the unfactored Unified Mask (TSM) achieves high appearance insertion AUC ($0.81$ EGTEA, $0.59$ EPIC) and $\text{F}^{+}$ ($0.99$, $0.99$) simply by retaining all frames and larger spatial footprints ($20.2\%$, $21.1\%$). Lacking temporal gating, it fails at precise localization, yielding worse appearance deletion AUCs than our factored approach ($0.43$ vs. $0.27$ EGTEA; $0.22$ vs. $0.1$ EPIC) and offering no motion explanation. This indicates that the dual-mask factorization, rather than the shared perturbation pipeline, drives the deletion gains. Appendix~\ref{sec:ablations} reports a cumulative component ablation, an ablation of the noun/verb semantic split (App.~\ref{app:semantic_split}), the flow-alignment ablation (App.~\ref{app:flow_ablation}), and hyperparameter sweeps (App.~\ref{app:sensitivity}). 

\paragraph{Auditing example.} Grouping EGTEA predictions by verb and ranking classes by few active $\mm$ frames and high confidence surfaces nominally temporal verbs whose explanations collapse to almost no frames: for \emph{wash}, $\mm$ keeps on average $0.8$ of $8$ frames while the classifier remains $\approx 90\%$ confident (App.~\ref{app:audit}). This flags \emph{wash} as a candidate static-appearance shortcut, a hypothesis that a practitioner can then test by intervention.

\paragraph{Cost.} Optimization takes $11.7$\,s per clip on TSM, comparable to STEP ($11.4$\,s) and EP ($10.3$\,s). Because mask parameters are independent across clips, batching eight clips reduces the amortized cost by about $6.5\times$, to roughly $2$\,s per clip (App.~\ref{app:compute}).
\section{Conclusions and Limitations}\label{sec:discussion}

We frame video action explanation as a disentangled two-mask problem, isolating spatial appearance and temporal dynamics. Our core mechanisms: factored perturbation, STE-based frame replacement, separation loss, and a one-sided ReLU area constraint are  individually motivated and jointly necessary to guarantee explanatory precision.  
Against STEP, Extremal Perturbation, and unified-mask baselines, our method consistently attains the lowest deletion AUC and highest complementary fidelity ($\text{F}^{-}$) on both axes, while producing masks roughly $5\times$ sparser than the strongest competitor. By separately highlighting exactly \emph{where} and \emph{when} a model looks, our factored explanations provide practitioners with a precise post-hoc auditing tool to expose spurious shortcuts. 
Our proposed framework serves strictly as a post-hoc auditing tool for frozen video classifiers, producing spatial and temporal masks that expose the model's internal reliance.   
Because it provides diagnostic evidence rather than inherent interpretability or accuracy improvements, practitioners should use it to surface candidate failure modes and shortcuts, avoiding the misuse of confident-looking saliency maps to arbitrarily validate incorrect predictions. 
However, the framework remains limited by its reliance on a flow alignment loss for single-head models, which introduces an inductive bias toward motion-rich regions. The factorization separates the two perturbation axes by construction, but it does not force the classifier's representations to separate; measured cross-axis gradients are substantial. Finally, factoring strictly by appearance and motion is inherently restrictive, making the method unsuitable for actions defined by audio cues, gaze patterns, off-frame interactions or long-range interactions that last more than 8 frames.

\begin{ack}
This work was funded by Oregon State University and its School of Electrical Engineering and Computer Science. We are grateful to Fuxin Li for valuable advice, and to Devika Venkatesh and Dmitriy Shmidt for their help with computational resources.
\end{ack}

\bibliographystyle{plainnat}
\bibliography{refs}

@inproceedings{lin2019tsm,
  title={TSM: Temporal Shift Module for Efficient Video Understanding},
  author={Lin, J. and Gan, C. and Han, S.},
  booktitle={Proceedings of the IEEE/CVF International Conference on Computer Vision (ICCV)},
  year={2019}
}

@inproceedings{damen2018scaling,
  title={Scaling egocentric vision: The EPIC-Kitchens dataset},
  author={Damen, D. and others},
  booktitle={Proceedings of the European Conference on Computer Vision (ECCV)},
  year={2018}
}

@inproceedings{li2018egtea,
  title={In the eye of beholder: Joint learning of gaze and actions in first person video},
  author={Li, Y. and Liu, M. and Rehg, J. M.},
  booktitle={Proceedings of the European Conference on Computer Vision (ECCV)},
  year={2018}
}

@inproceedings{goyal2017something,
  title={The ``Something Something'' video database for learning and evaluating visual common sense},
  author={Goyal, R. and others},
  booktitle={Proceedings of the IEEE International Conference on Computer Vision (ICCV)},
  year={2017}
}

@inproceedings{feichtenhofer2019slowfast,
  title={SlowFast networks for video recognition},
  author={Feichtenhofer, C. and Fan, H. and Malik, J. and He, K.},
  booktitle={Proceedings of the IEEE/CVF International Conference on Computer Vision (ICCV)},
  year={2019}
}

@inproceedings{liu2022videoswin,
  title={Video Swin Transformer},
  author={Liu, Z. and others},
  booktitle={Proceedings of the IEEE/CVF Conference on Computer Vision and Pattern Recognition (CVPR)},
  year={2022}
}

@inproceedings{selvaraju2017grad,
  title={Grad-CAM: Visual explanations from deep networks via gradient-based localization},
  author={Selvaraju, R. R. and others},
  booktitle={Proceedings of the IEEE International Conference on Computer Vision (ICCV)},
  year={2017}
}

@inproceedings{petsiuk2018rise,
  title={RISE: Randomized input sampling for explanation of black-box models},
  author={Petsiuk, V. and Das, A. and Saenko, K.},
  booktitle={Proceedings of the British Machine Vision Conference (BMVC)},
  year={2018}
}

@inproceedings{wang2020score,
  title={Score-CAM: Score-weighted visual explanations for convolutional neural networks},
  author={Wang, H. and others},
  booktitle={Proceedings of the IEEE/CVF Conference on Computer Vision and Pattern Recognition (CVPR) Workshops},
  year={2020}
}

@inproceedings{li2021towards,
  title={Towards visually explaining video understanding networks with perturbation},
  author={Li, Z. and others},
  booktitle={Proceedings of the IEEE/CVF Winter Conference on Applications of Computer Vision (WACV)},
  year={2021}
}

@inproceedings{fong2019extremal,
  title={Understanding deep networks via extremal perturbations and smooth masks},
  author={Fong, R. and Patrick, M. and Vedaldi, A.},
  booktitle={Proceedings of the IEEE/CVF International Conference on Computer Vision (ICCV)},
  year={2019}
}

@inproceedings{stergiou2019saliency,
  title={Saliency tubes: Visual explanations for spatio-temporal convolutions},
  author={Stergiou, A. and Kapidis, G. and Kalliatakis, G. and Chrysoulas, C. and Veltkamp, R. and Poppe, R.},
  booktitle={Proceedings of the IEEE International Conference on Image Processing (ICIP)},
  year={2019}
}

@inproceedings{hartley2022swagv,
  title={SWAG-V: Explanations for video using superpixels weighted by average gradients},
  author={Hartley, T. and Sidorov, K. and Willis, C. and Marshall, D.},
  booktitle={Proceedings of the IEEE/CVF Winter Conference on Applications of Computer Vision (WACV)},
  year={2022}
}

@inproceedings{ji2023spatiotemporal,
  title={Spatial-temporal concept based explanation of 3D ConvNets},
  author={Ji, Y. and Wang, Y. and Kato, J.},
  booktitle={Proceedings of the IEEE/CVF Conference on Computer Vision and Pattern Recognition (CVPR)},
  year={2023}
}

@inproceedings{kowal2024vtcd,
  title={Understanding video transformers via universal concept discovery},
  author={Kowal, M. and Dave, A. and Ambrus, R. and Gaidon, A. and Derpanis, K. G. and Tokmakov, P.},
  booktitle={Proceedings of the IEEE/CVF Conference on Computer Vision and Pattern Recognition (CVPR)},
  year={2024}
}

@inproceedings{saha2024exploring,
  title={Exploring explainability in video action recognition},
  author={Saha, A. and Gupta, S. and Ankireddy, S. K. and Chahine, K. and Ghosh, J.},
  booktitle={Proceedings of the IEEE/CVF Conference on Computer Vision and Pattern Recognition (CVPR) Workshops},
  year={2024}
}

@article{twinanda2016endonet,
  title={Endonet: a deep architecture for recognition tasks on laparoscopic videos},
  author={Twinanda, Andru P and Shehata, Sherif and Mutter, Didier and Marescaux, Jacques and De Mathelin, Michel and Padoy, Nicolas},
  journal={IEEE transactions on medical imaging},
  volume={36},
  number={1},
  pages={86--97},
  year={2016},
  publisher={IEEE}
}

@article{funke2019video,
  title={Video-based surgical skill assessment using 3D convolutional neural networks},
  author={Funke, Isabel and Mees, S{\"o}ren Torge and Weitz, J{\"u}rgen and Speidel, Stefanie},
  journal={International journal of computer assisted radiology and surgery},
  volume={14},
  number={7},
  pages={1217--1225},
  year={2019},
  publisher={Springer}
}

@article{knab2025concepts,
  title={Concepts in Motion: Temporal Concept Bottleneck Model for Interpretable Video Classification},
  author={Knab, Patrick and Marton, Sascha and Schubert, Philipp J and Guggiana, Drago and Bartelt, Christian},
  journal={arXiv preprint arXiv:2509.20899},
  year={2025}
}

@inproceedings{Lee2025,
  title={Disentangled concepts speak louder than words: Explainable video action recognition},
  author={Lee, J. and Lee, W. and Park, G.-M. and Kim, S. T. and Choi, J.},
  booktitle={Advances in Neural Information Processing Systems (NeurIPS)},
  year={2025}
}

@article{jang2016categorical,
  title={Categorical reparameterization with gumbel-softmax},
  author={Jang, Eric and Gu, Shixiang and Poole, Ben},
  journal={arXiv preprint arXiv:1611.01144},
  year={2016}
}

@inproceedings{gberta_2021_ICML,
    author  = {Gedas Bertasius and Heng Wang and Lorenzo Torresani},
    title = {Is Space-Time Attention All You Need for Video Understanding?},
    booktitle   = {Proceedings of the International Conference on Machine Learning (ICML)}, 
    month = {July},
    year = {2021}
}

@inproceedings{zhuo2019explainable,
  title={Explainable video action reasoning via prior knowledge and state transitions},
  author={Zhuo, T. and Cheng, Z. and Zhang, P. and Wong, Y. and Kankanhalli, M.},
  booktitle={Proceedings of the ACM International Conference on Multimedia},
  year={2019}
}

@inproceedings{farneback2003two,
  title={Two-frame motion estimation based on polynomial expansion},
  author={Farneb{\"a}ck, G.},
  booktitle={Proceedings of the Scandinavian Conference on Image Analysis},
  year={2003}
}
\appendix

\newpage
\section{Appendix}

\subsection{Reproducibility statement}
Code, the hyperparameter search ranges and selected values (App.~\ref{app:hyperparams}), the evaluation clip lists, and the TimeSformer checkpoints trained on EPIC-Kitchens-55 and EGTEA Gaze+ are available at \url{https://github.com/vbhavank/wherewhen}. Frozen SSv2 weights are taken from the official releases of the cited backbones. All results are supported by paired significance tests (App.~\ref{app:stats}). Like the classifiers it explains, the method could be used to refine action recognition systems deployed for surveillance; we release it only as an analysis tool for existing models.

\setcounter{proposition}{0}
\subsection{Qualitative Results}

\begin{figure}[h!]
    \centering
    \includegraphics[width=\linewidth]{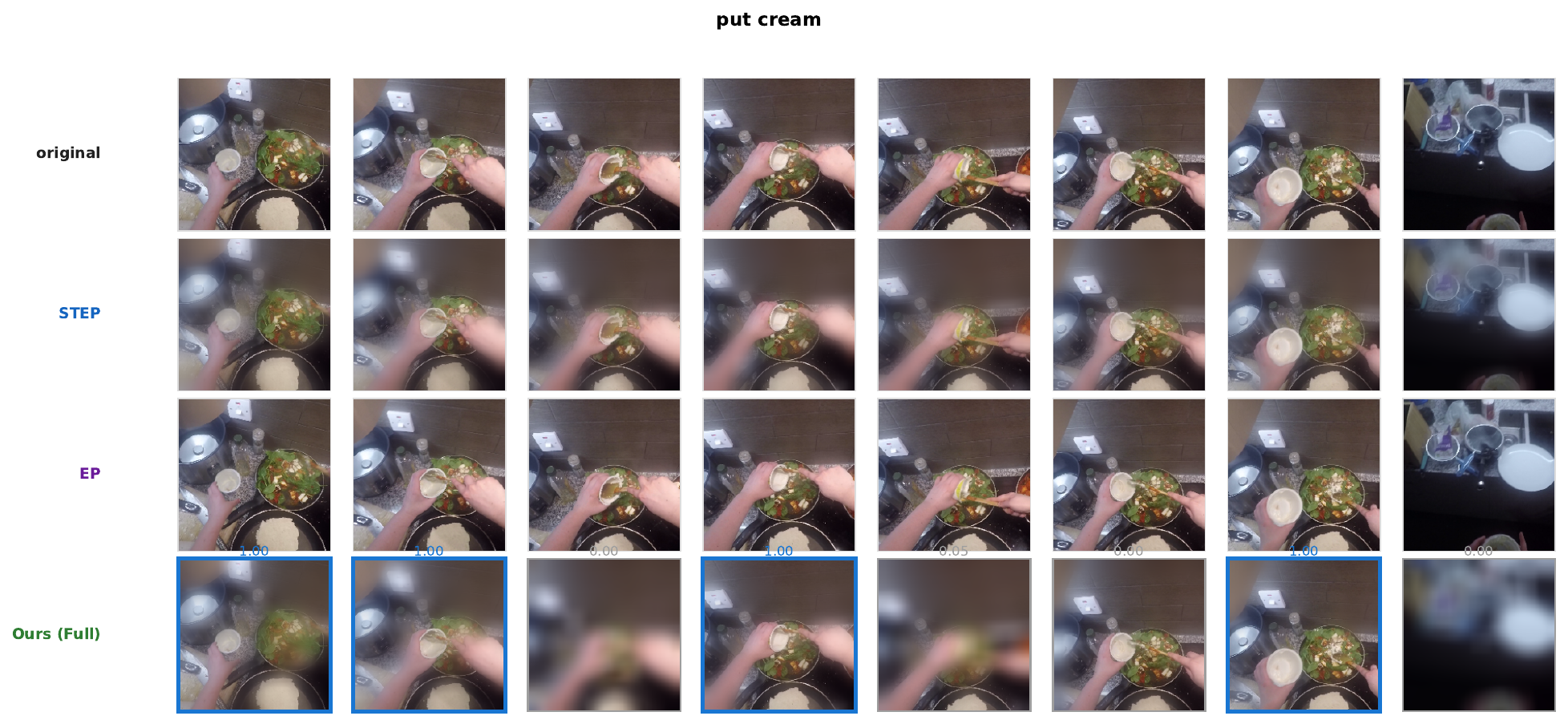}
    \includegraphics[width=\linewidth]{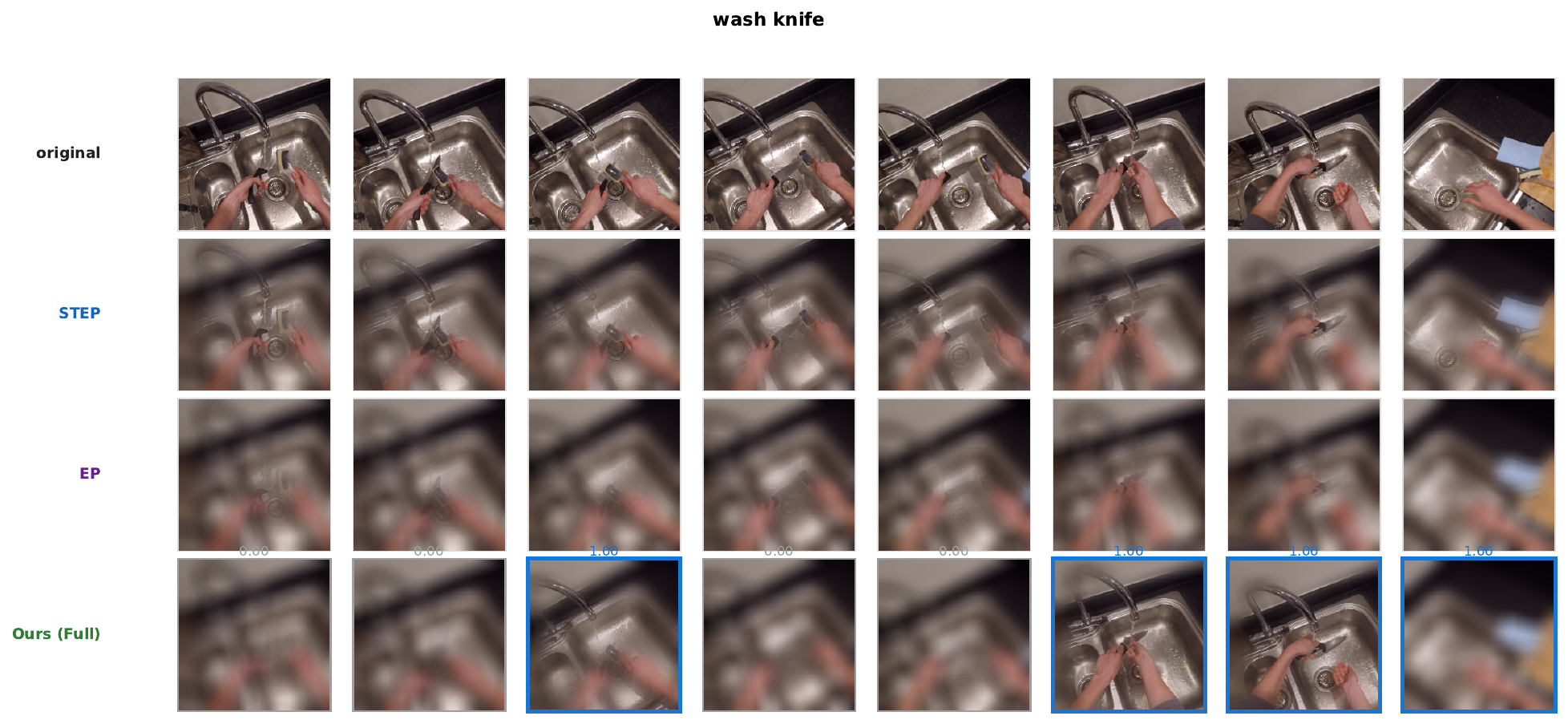}
    \caption{Qualitative comparison of explanation methods on a single Epic-Kitchens clip for action \textit{put-cream} and \textit{wash-knife}. Each column is one of the 8 uniformly sampled frames. \textbf{Row 1 (original):} input frames. \textbf{Row 2 (STEP~\cite{li2021towards}):} spatial mask $M_a$ learned by STEP; the highlighted region is shown sharp while the complement is replaced by a Gaussian blur ($31{\times}31$, $\sigma{=}20$), matching the perturbation seen by the model. \textbf{Row 3 (Extremal Perturbation~\cite{fong2019extremal}):} same rendering for the EP spatial mask. \textbf{Row 4 (Ours):} our disentangled spatial mask under the same blur-blend rendering; frame borders are \textcolor{blue}{blue} when the temporal mask $M_m{>}0.5$ (model-relevant frame) and \textcolor{gray}{gray} when frozen (nearest-active-frame replacement applied). Values above each frame give the soft $M_m$ weight.}
    \label{fig:qual3}
\end{figure}

\subsection{Ablations}\label{sec:ablations}
\begin{figure}[h!]
    \centering
    \includegraphics[height=0.32\linewidth, width=\linewidth]{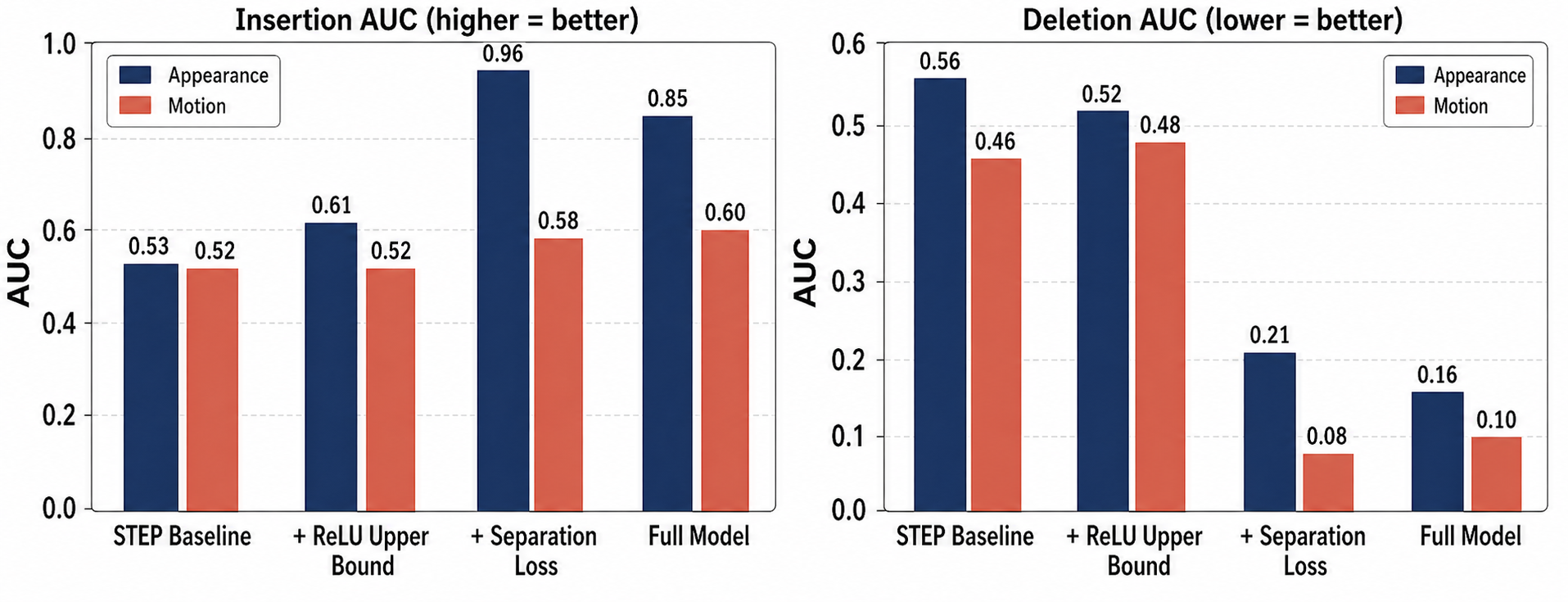}
    \caption{Cumulative ablation on a frozen TSM trained on EGTEA Gaze+. Each bar adds a component on top of the previous configuration.}
    \label{fig:ablation}
\end{figure}

We cumulatively ablate our framework from the STEP baseline on EGTEA (frozen TSM; Figure~\ref{fig:ablation}). First, replacing MSE with the ReLU upper-bound area loss prevents non-discriminative masking by allowing sub-budget use, improving Appearance Insertion AUC (0.53$\to$0.61) and Deletion AUC (0.56$\to$0.52). Next, adding separation loss ($\lsep$) prevents spatial masking on frozen frames, driving massive faithfulness gains: Appearance Deletion AUC drops to 0.21, Motion Deletion drops drastically from 0.48 to 0.08, and Appearance Insertion increases to 0.96. Finally, the Full Model (incorporating STE hard frame selection) solidifies temporal perturbations, achieving our best Appearance Deletion (0.16) and peak Motion Insertion (0.60), while Motion Deletion remains highly competitive (0.10) and Appearance Insertion normalizes to 0.85.

\section{Theoretical analysis}\label{app:theory}

This appendix formalizes four claims from the main text. Section~\ref{app:thm:ste} bounds the bias of the Straight-Through Estimator used for hard frame selection (\S\ref{sec:perturbation}). Section~\ref{app:thm:area} characterizes the stationary points of the symmetric MSE area loss and the proposed ReLU upper-bound area loss (\S\ref{sec:losses}), formalizing why the latter permits sub-budget masks.  Section~\ref{app:thm:ins_del} formalizes the structural advantage of deletion AUC over insertion AUC for sparse explanations (\S\ref{sec:experiments}).

We adopt the following notation. Vectors and tensors are bold; scalar entries are non-bold and indexed by subscripts. The classifier $f$ is twice continuously differentiable as a function of its input, which holds for all architectures considered in the main text. The clean clip is $\xrgb \in \R^{T \times H \times W \times 3}$, the spatially-perturbed clip is $\mathbf{x}^{\text{app}}$ as in \S\ref{sec:perturbation}, and the final perturbed clip is $\xhat$ as in Eq.~\eqref{eq:full_perturbation}. We write $\nabla_\theta \mathcal{L}$ for the gradient of a scalar loss with respect to a parameter tensor $\theta$, and $\sg(\cdot)$ for the stop-gradient operator (zero in the backward pass, identity in the forward pass).

\subsection{Bias of the Straight-Through Estimator for hard frame selection}\label{app:thm:ste}

The frame replacement scheme in \S\ref{sec:perturbation} employs the gradient estimator
\begin{equation}\label{eq:ste_estimator}
\tilde{w}_{i \to j} \;=\; \mathbbm{1}[j = j^*(i)] \;-\; \sg(w_{i \to j}) \;+\; w_{i \to j},
\end{equation}
where $j^*(i) = \arg\max_j w_{i \to j}$ and $w_{i \to j}$ is the soft replacement weight defined in \S\ref{sec:perturbation}. The forward pass of Eq.~\eqref{eq:ste_estimator} evaluates to the one-hot vector $\mathbf{e}_{j^*(i)}$, while the backward pass yields the gradient of the soft weights:
\[
\frac{\partial \tilde{w}_{i \to j}}{\partial m_k} \;=\; \frac{\partial w_{i \to j}}{\partial m_k}.
\]
For any downstream scalar loss $\mathcal{L}$ dependent on the perturbed clip, the STE backward computation produces
\begin{equation}\label{eq:ste_backward}
\widehat{\nabla_{m_k} \mathcal{L}}
\;=\; \sum_{i,j} \frac{\partial \mathcal{L}}{\partial \tilde{w}_{i \to j}} \cdot \frac{\partial w_{i \to j}}{\partial m_k},
\end{equation}
whereas the gradient of the true forward function (one-hot $\arg\max$) is zero almost everywhere and undefined on a measure-zero set of ties. Following \cite{jang2016categorical}, the relevant comparison is therefore not against the ``true'' gradient, which provides no signal, but against the gradient of a surrogate: the loss evaluated on the fully soft mixture.
\paragraph{Setup.} Let $\xhat_{\text{soft}}(\mathbf{m})$ denote the perturbed clip computed using soft weights $w$ instead of the STE-corrected $\tilde{w}$, and let $\mathcal{L}_{\text{soft}}(\mathbf{m}) := \mathcal{L}(\xhat_{\text{soft}}(\mathbf{m}))$. The STE gradient~\eqref{eq:ste_backward} equals $\nabla_{\mathbf{m}} \mathcal{L}_{\text{soft}}$ evaluated through $\xhat_{\text{soft}}$, but in our pipeline the loss is evaluated at $\xhat = \xhat_{\text{hard}}$, the discretized clip. The bias is formalized as
\begin{equation}\label{eq:ste_bias}
\text{Bias}(\mathbf{m})
\;=\; \nabla_{\mathbf{m}} \mathcal{L}_{\text{soft}}\big|_{\xhat = \xhat_{\text{hard}}}
\;-\; \nabla_{\mathbf{m}} \mathcal{L}_{\text{soft}}\big|_{\xhat = \xhat_{\text{soft}}}.
\end{equation}

Let $L$ be a Lipschitz constant of $\nabla \mathcal{L}_{\text{soft}}$ with respect to $\xhat$ under the $\ell_1$ norm, viewed as a function from clips to gradients with respect to $\mathbf{m}$. Such a constant exists for any twice continuously differentiable classifier on a bounded input domain.

 Intuitively, this bounds the deviation of the STE gradient from that of a strictly soft pipeline. The bias is minimized when the soft argmax distribution is highly peaked (confident) or when the candidate replacement frames are visually similar. 

\begin{proposition}[STE bias bound]\label{prop:ste_bias}
For each frame index $i$, let $\Delta_i := w_{i \to j^*(i)} - w_{i \to j^{(2)}(i)}$ denote the gap between the largest and second-largest replacement weight, and let $\delta_i := \max_{j \neq j^*(i)} \|\mathbf{x}^{\textnormal{app}}_{j^*(i)} - \mathbf{x}^{\textnormal{app}}_j\|$ denote the largest pairwise distance between candidate replacement frames. Then
\begin{equation}\label{eq:ste_bound}
\big\|\textnormal{Bias}(\mathbf{m})\big\|
\;\leq\; L \cdot \sum_{i \,:\, \mm^{\textnormal{STE}}_i = 0}
(1 - \Delta_i) \cdot \delta_i.
\end{equation}
\end{proposition}

\begin{proof}
Fix a frame $i$ marked frozen by the temporal STE gate ($\mm^{\text{STE}}_i = 0$, replacement active). The hard and soft replacement frames at position $i$ are
\[
\xhat^{\text{hard}}_i = \mathbf{x}^{\text{app}}_{j^*(i)},
\qquad
\xhat^{\text{soft}}_i = \sum_j w_{i \to j} \, \mathbf{x}^{\text{app}}_j.
\]

We need to measure the visual distance between the ``hard'' frame (if we forced the model to pick exactly one frame) and the ``soft'' blended frame (the weighted average the math thinks we chose).

First, we subtract the soft frame from the hard frame. Because the soft frame is just a mixture of the winning frame and all the losing frames, the winning frame perfectly cancels out its own portion of the blend. We are left with just a sum of all the ``losing'' frames ($j \neq j^*$), weighted by their probabilities ($w_{i \to j}$), multiplied by how visually different they are from the winner:
\[
    \|\xhat^{\text{hard}}_i - \xhat^{\text{soft}}_i\| 
    \;=\; \Big\|\sum_{j \neq j^*(i)} w_{i \to j} \big(\mathbf{x}^{\text{app}}_{j^*(i)} - \mathbf{x}^{\text{app}}_j\big)\Big\|
\]

Calculating the exact visual difference for every single losing frame is too complex, so we establish an upper bound. The Triangle Inequality allows us to pull the absolute value (norm) inside the summation. Instead of tracking exact differences, we assume the worst-case scenario. Let $\delta_i$ represent the absolute biggest visual difference between the winning frame and \textit{any} candidate frame. By replacing the exact distances with this maximum distance, we guarantee our formula acts as a strict ceiling ($\leq$):
\[
    \Big\|\sum_{j \neq j^*(i)} w_{i \to j} \big(\mathbf{x}^{\text{app}}_{j^*(i)} - \mathbf{x}^{\text{app}}_j\big)\Big\| 
    \;\leq\; \sum_{j \neq j^*(i)} w_{i \to j} \cdot \delta_i
\]

Because the weights ($w$) are probabilities, the sum of the probabilities for every candidate frame must equal $1$ ($100\%$). Therefore, the combined probability of all the \textit{losing} frames is simply $1$ minus the probability of the \textit{winning} frame ($w_{i \to j^*(i)}$). This neatly collapses the entire summation:
\[
    \sum_{j \neq j^*(i)} w_{i \to j} \cdot \delta_i 
    \;=\; \big(1 - w_{i \to j^*(i)}\big) \cdot \delta_i
\]

We define the gap ($\Delta_i$) as the exact difference in probability between the first-place winner and the second-place runner-up ($j^{(2)}$):
\[
    w_{i \to j^*(i)} \;=\; w_{i \to j^{(2)}(i)} + \Delta_i
\]
Because a probability can never be negative ($w_{i \to j^{(2)}(i)} \geq 0$), the winning probability must be \textit{at least} as large as the gap itself: $w_{i \to j^*(i)} \geq \Delta_i$. 

If the winning probability is greater than or equal to $\Delta_i$, then subtracting it from $1$ flips the inequality sign, giving us:
\[
    1 - w_{i \to j^*(i)} \;\leq\; 1 - \Delta_i
\]

By substituting this gap logic back into our collapsed equation from, we arrive at the final pointwise bound. The visual error between the hard choice and the soft choice is mathematically guaranteed to be less than or equal to the visual difference of the worst frame ($\delta_i$) multiplied by the uncertainty of the decision ($1 - \Delta_i$):
\[
    \|\xhat^{\text{hard}}_i - \xhat^{\text{soft}}_i\| \;\leq\; (1 - \Delta_i) \cdot \delta_i.
\]

Summing the norms over all frozen frames bounds the $\ell_1$ distance between the hard and soft clips. Applying the $\ell_1$-Lipschitz continuity of $\nabla \mathcal{L}_{\text{soft}}$ then yields Eq.~\eqref{eq:ste_bound}.
\end{proof}

\paragraph{Analysis of the Bias Bound.} The bound comprises two factors per frozen frame, capturing independent properties of the replacement scheme. The factor $(1 - \Delta_i)$ quantifies the entropy of the soft argmax; a confident replacement (i.e., a large margin between the top two weights) minimizes this term. The factor $\delta_i$ measures the visual discrepancy among candidate replacement frames; for temporally redundant segments, $\delta_i$ is small, rendering the specific replacement choice inconsequential.

These factors interact with the temperature annealing schedule and replacement geometry. Because temperature annealing is applied to the \emph{mask} sigmoid rather than the replacement weights directly, it does not force $\Delta_i \to 1$ asymptotically. Instead, it pushes the mask values toward $\{0,1\}$, yielding an exactly one-hot $\tilde{w}$ during the forward pass and restricting the active set to a sparse subset of frames. Within this active set, the proximity Gaussian $\exp(-\|i - j\|^2 / 2\sigma_R^2)$ (where $\sigma_R = 2$) separates the nearest active neighbor from the second nearest by a non-trivial multiplicative factor (e.g., $\exp(-1/8)/\exp(-1/2) \approx 1.46$ for distances $1$ and $2$). Consequently, $\Delta_i$ is lower-bounded by a constant determined solely by the active set's geometry. Furthermore, $\delta_i$ remains small when the nearest-active-frame replacement selects a visually similar source, which is typical for slowly varying video segments. 

\subsection{Stationary points of the area constraints}\label{app:thm:area}
In our method, we want the neural network to highlight 
important parts of a video. We give it a "budget" (a target size for the highlight mask, called $v$). We want to prove mathematically why using a standard Mean Squared Error (MSE) loss forces the mask to match this budget \textit{exactly}, whereas using a ReLU loss allows the mask to be \textit{smaller} than the budget if it finds the important features early. Let us characterize the stationary points of the joint loss $\mathcal{L}_{\text{pred}}(\mathbf{m}) + \lambda \mathcal{L}_{\text{area}}(\mathbf{m}, v)$ under both the symmetric MSE and the proposed ReLU upper-bound area constraints.

\paragraph{Setup.} Let $\mathbf{m} \in [0,1]^N$ denote a generic mask (representing $\ma$ flattened, or $\mm$, or any sub-block thereof). Let $\mathcal{L}_{\text{pred}}: [0,1]^N \to \R_{\geq 0}$ be continuously differentiable with bounded gradients, $\|\nabla \mathcal{L}_{\text{pred}}(\mathbf{m})\| \leq L$ for some constant $L$. 
In practice $\mathcal{L}_{\text{pred}}$ has plateau regions where $\nabla \mathcal{L}_{\text{pred}}(\mathbf{m}^*) = \mathbf{0}$ once the mask covers enough of the model's evidence.

Let $v \in (0,1)$ denote the target area fraction and $t_i \in \{0, 1\}$ the binary target with $t_i = 1$ for $i \geq (1-v)N$ in sorted order. We write $\mathbf{m}^{\text{sort}}$ for the descending sort of $\mathbf{m}$ and abbreviate $a := \|\mathbf{m}\|_1 / N$ for the realized area fraction.

The symmetric and one-sided area losses are
\begin{equation}
\mathcal{L}_{\text{area}}^{\text{MSE}}(\mathbf{m}, v)
\;=\; \frac{1}{N} \sum_{i=1}^N \big(\mathbf{m}^{\text{sort}}_i - t_i\big)^2,
\qquad
\mathcal{L}_{\text{area}}^{\text{ReLU}}(\mathbf{m}, v)
\;=\; \frac{1}{N} \sum_{i=1}^N \big[\text{ReLU}(\mathbf{m}^{\text{sort}}_i - t_i)\big]^2.
\end{equation}

The symmetric area loss formulation invariably produces masks that saturate the allocated budget. As the penalty weight $\lambda$ increases, the realized mask area converges to the target budget at a rate of $O(1/\lambda)$.

\begin{proposition}[MSE forces budget saturation, with quantitative rate]\label{prop:mse_saturation}
Let $\mathbf{m}^*_\lambda$ be an interior stationary point of $\mathcal{L}_{\textnormal{pred}} + \lambda \mathcal{L}_{\textnormal{area}}^{\textnormal{MSE}}$. Then the realized area satisfies
\begin{equation}\label{eq:mse_rate}
|a^*_\lambda - v| \;\leq\; \frac{L \sqrt{N}}{2\lambda}.
\end{equation}
In particular, $a^*_\lambda \to v$ as $\lambda \to \infty$, and the gap closes at rate $O(1/\lambda)$ (with constant scaling as $\sqrt{N}$).
\end{proposition}

\begin{proof}
At an interior stationary point, the first-order condition dictates that the total gradient must equal zero:
\[
\nabla \mathcal{L}_{\text{pred}}(\mathbf{m}^*_\lambda) + \lambda \nabla \mathcal{L}_{\text{area}}^{\text{MSE}}(\mathbf{m}^*_\lambda, v) = \mathbf{0}.
\]
Rearranging this equivalence yields $\lambda \nabla \mathcal{L}_{\text{area}}^{\text{MSE}}(\mathbf{m}^*_\lambda, v) = -\nabla \mathcal{L}_{\text{pred}}(\mathbf{m}^*_\lambda)$. By taking the norm of both sides and applying the assumed gradient bound $\|\nabla \mathcal{L}_{\text{pred}}\| \leq L$, we deduce that the norm of the area loss gradient must be strictly bounded: 
\[
\|\nabla \mathcal{L}_{\text{area}}^{\text{MSE}}(\mathbf{m}^*_\lambda, v)\| \leq \frac{L}{\lambda}.
\]
This mathematically enforces that as the penalty weight $\lambda$ grows, the area loss gradient is driven toward zero. To evaluate this gradient norm, we compute the partial derivative of the MSE area loss with respect to a specific mask parameter $m_k$. Applying the chain rule to the quadratic penalty yields $\partial \mathcal{L}_{\text{area}}^{\text{MSE}} / \partial m_k = (2/N)(\mathbf{m}^{\text{sort}}_{i_k} - t_{i_k})$, where $i_k$ represents the rank index of position $k$ after the descending sort. 

Computing the squared Euclidean norm ($\ell_2$ norm) of this gradient vector entails summing the squares of its components over all $N$ dimensions:
\[
\|\nabla \mathcal{L}_{\text{area}}^{\text{MSE}}\|_2^2
\;=\; \frac{4}{N^2} \sum_{i=1}^N (\mathbf{m}^{\text{sort}}_i - t_i)^2
\;\leq\; \frac{L^2}{\lambda^2}.
\]
Equating this algebraic expansion to our previously established upper bound $(L/\lambda)^2$ and isolating the summation yields a bound on the sum of squared differences: $\sum_i (\mathbf{m}^{\text{sort}}_i - t_i)^2 \leq N^2 L^2 / (4\lambda^2)$. 

We now relate this squared Euclidean distance to the absolute difference in total area. By invoking the Cauchy--Schwarz inequality, which asserts that the sum of elements ($\ell_1$ norm) is bounded by the square root of the sum of their squares ($\ell_2$ norm) scaled by the dimension $N$, we obtain:
\[
\Big| \sum_i (\mathbf{m}^{\text{sort}}_i - t_i) \Big|
\;\leq\; \sqrt{N \cdot \sum_i (\mathbf{m}^{\text{sort}}_i - t_i)^2}
\;\leq\; \frac{N L \sqrt{N}}{2\lambda}.
\]
To conclude the proof, we recognize that the realized area $a^*_\lambda$ is simply the mean of the sorted mask $(1/N) \sum_i \mathbf{m}^{\text{sort}}_i$, and the target budget $v$ is the mean of the binary targets $(1/N) \sum_i t_i$.
\begin{align*}
    a^*_\lambda &= \frac{1}{N} \sum_{i=1}^N \mathbf{m}^{\text{sort}}_i \\
    v &= \frac{1}{N} \sum_{i=1}^N t_i
\end{align*}
We want to find the absolute difference between these two quantities. Substituting our definitions gives:
\[
    |a^*_\lambda - v| \;=\; \left| \left( \frac{1}{N} \sum_{i=1}^N \mathbf{m}^{\text{sort}}_i \right) - \left( \frac{1}{N} \sum_{i=1}^N t_i \right) \right|
\]
Because $\frac{1}{N}$ is a positive constant shared by both terms, we can factor it completely outside of the absolute value, combining the summations into a single term:
\[
    |a^*_\lambda - v| \;=\; \frac{1}{N} \left| \sum_{i=1}^N (\mathbf{m}^{\text{sort}}_i - t_i) \right|
\]
From our earlier application of the Cauchy-Schwarz inequality, we established a strict upper bound for this absolute sum: $\big| \sum_{i=1}^N (\mathbf{m}^{\text{sort}}_i - t_i) \big| \leq \frac{N L \sqrt{N}}{2\lambda}$. We substitute this upper bound into our equation:
\[
    |a^*_\lambda - v| \;\leq\; \frac{1}{N} \cdot \left( \frac{N L \sqrt{N}}{2\lambda} \right)
\]
The $N$ in the denominator perfectly cancels the $N$ in the numerator of the bound, leaving us withEq.~\eqref{eq:mse_rate}:
\[
    |a^*_\lambda - v| \;\leq\; \frac{L \sqrt{N}}{2\lambda}
\]
\end{proof}

Conversely, the one-sided area loss permits the mask to converge below the target budget, provided the prediction loss is minimized. Once the mask area strictly satisfies the budget constraint, the area penalty yields a zero gradient, thereby allowing the realized sparsity to be dictated entirely by the prediction loss landscape.

\begin{proposition}[ReLU permits sub-budget stationary points independent of $\lambda$]\label{prop:relu_subbudget}
Suppose there exists $\mathbf{m}^* \in [0,1]^N$ with $\nabla \mathcal{L}_{\textnormal{pred}}(\mathbf{m}^*) = \mathbf{0}$ and $\mathbf{m}^{\textnormal{sort},*}_i \leq t_i$ for all $i$ (so $a^* \leq v$). Then $\mathbf{m}^*$ is a stationary point of $\mathcal{L}_{\textnormal{pred}} + \lambda \mathcal{L}_{\textnormal{area}}^{\textnormal{ReLU}}$ for every $\lambda > 0$.
\end{proposition}

\begin{proof}
We analyze the stationary condition by first computing the partial derivative of the ReLU area loss. Applying the chain rule to the squared ReLU function gives:
\[
\frac{\partial \mathcal{L}_{\text{area}}^{\text{ReLU}}}{\partial m_k}
\;=\; \frac{2}{N}\,\text{ReLU}\big(\mathbf{m}^{\text{sort}}_{i_k} - t_{i_k}\big).
\]
The critical property of the ReLU function is that it evaluates to zero for any non-positive argument. Under the hypothesis that the proposed mask strictly satisfies the budget constraint (i.e., $\mathbf{m}^{\text{sort},*}_i \leq t_i$ for all $i$), the argument $(\mathbf{m}^{\text{sort},*}_i - t_{i_k})$ is inherently non-positive across all dimensions. Consequently, the ReLU argument returns zero, and the gradient evaluates to exactly zero everywhere.

Substituting this zero-gradient result back into the global first-order stationary condition yields:
\[
\nabla \mathcal{L}_{\text{pred}}(\mathbf{m}^*) + \lambda \cdot \mathbf{0} = \mathbf{0}.
\]
Because the area penalty gradient vanishes completely, the optimality of the point relies solely on the prediction loss landscape. Since we posited by hypothesis that $\nabla \mathcal{L}_{\text{pred}}(\mathbf{m}^*) = \mathbf{0}$, this equality holds true independent of the magnitude of the hyperparameter $\lambda$.
\end{proof}

\paragraph{What the two propositions say jointly.} Proposition~\ref{prop:mse_saturation} formalizes the empirical observation that MSE-area-loss masks always realize close to the budget. The symmetric penalty penalizes the mask for being smaller than $v$ as well as for being larger, so any interior stationary point is pushed toward $a = v$ at rate $O(1/\lambda)$. Proposition~\ref{prop:relu_subbudget} shows that the ReLU form decouples realized area from budget: once the mask is sub-budget, the area term contributes nothing to the gradient, and the realized sparsity is determined entirely by the geometry of $\mathcal{L}_{\text{pred}}$. This is the behavior we exploit in Table~\ref{tab:dataset_model_results}, where realized appearance-mask sparsity falls well below the budget set during optimization.

\subsection{Insertion AUC has a structural advantage for larger masks}\label{app:thm:ins_del}

We now formalize the analytical claim introduced in Section~\ref{sec:experiments}:  for explanation masks whose values are driven to $0$ or $1$, the Insertion AUC metric implicitly rewards mask size, whereas the Deletion AUC metric strictly resists this confounding factor. The effect is specific to such (near-)binary masks and does not arise in the same form for continuous attribution maps such as saliency maps. Perturbation methods that optimize masks toward binary values (our method, Extremal Perturbation and STEP, all of which anneal or threshold their masks toward $\{0,1\}$) fall in the binary regime analyzed here.

\paragraph{Setup.} Fix an input clip and a designated target classification class. For a given explanatory mask $M$ evaluated over the $N := THW$ spatial-temporal positions, the standard insertion procedure sequentially reveals the top-$\lfloor \alpha N \rfloor$ positions in descending order of mask magnitude (utilizing consistent permutation for tie-breaking). All remaining unrevealed positions are subjected to a baseline blur, and the target-class probability is subsequently evaluated. We denote this resulting evaluation curve as $C^{(M)}: [0,1] \to [0,1]$ and the normalized area-fraction of the active mask as $a(M) := |\{i : M_i = 1\}|/N$. By definition, the boundary conditions evaluate to $C^{(M)}(0) = C_{\text{base}}$ (the fully blurred baseline prediction) and $C^{(M)}(1) = C_{\text{full}}$ (the original unperturbed prediction), both of which are strictly independent of the chosen mask $M$.

We comparatively analyze two distinct masks, $M^{(1)}$ and $M^{(2)}$, characterized by active area fractions $0 < a_1 < a_2 < 1$. We stipulate that both masks share an underlying \emph{position importance ordering} $\pi$. This implies their active sets are strictly nested: method $1$ isolates the top $a_1 N$ positions, while method $2$ isolates the top $a_2 N$ positions, both following the exact $\pi$-order. Furthermore, tie-breaking among all inactive positions ($M_i = 0$) relies on the identical permutation for both methods. This specific configuration mathematically models the scenario where method $2$'s supplementary active positions are precisely the marginal features that method $1$ evaluated as minimally important and subsequently excluded.

Definition~\ref{def:faithful} formalizes the intuitive premise that a mask captures the entirety of the model's predictive evidence and nothing extraneous. Condition (iii) represents the core substantive constraint: spatial-temporal positions excluded from the mask yield strictly zero informational value to the classifier until the complete input tensor is restored.

\begin{definition}[Faithfulness]\label{def:faithful}
A mask $M$ with corresponding insertion curve $C^{(M)}$ is defined as faithful if it satisfies the following axioms:
\begin{enumerate}[leftmargin=*, label=(\roman*)]
\item $C^{(M)}$ is strictly non-decreasing across the domain $[0,1]$ (revealing additional visual information never deteriorates the prediction confidence);
\item $C^{(M)}$ is concave over the domain $[0, a(M)]$ (each newly revealed active position contributes a weakly lesser increase in confidence than the preceding position);
\item $C^{(M)}(\alpha) = C^{(M)}(a(M))$ for all $\alpha \in [a(M), 1)$, with the original confidence $C^{(M)}(1) = C_{\textnormal{full}}$ recovered exclusively as a discontinuous jump at $\alpha = 1$.
\end{enumerate}
\end{definition}

Condition (iii) is the primary engine of the faithfulness assumption: positions deemed inactive by the mask contain zero predictive signal until the full context is forcibly restored. This constraint is mathematically stronger than the generic inequalities occasionally utilized in statements of this metric, allowing for rigorous geometric bounds in the ensuing proofs. While empirical explanation methods rarely satisfy Condition (iii) perfectly, we address the practical implications of this relaxation below.

\paragraph{Notation.} Let us define the baseline-adjusted confidence curves as $S(\alpha) := C^{(1)}(\alpha) - C_{\text{base}}$ and $T(\alpha) := C^{(2)}(\alpha) - C_{\text{base}}$. Both functions are non-negative and non-decreasing, anchored at $S(0) = T(0) = 0$ and terminating at $S(1) = T(1) = C_{\text{full}} - C_{\text{base}}$. For the fractional domain $\alpha \leq a_1$, the insertion procedures of both methods reveal mathematically identical spatial-temporal coordinates (because their importance orderings are shared, and method $1$'s active set acts as a strict prefix to method $2$'s). Consequently, $S(\alpha) = T(\alpha)$ precisely on $[0, a_1]$. We denote the pointwise deviation between the curves as $\Delta(\alpha) := T(\alpha) - S(\alpha) \geq 0$.

\begin{proposition}[Insertion AUC favors larger faithful masks]\label{prop:ins_advantage}
Assuming the setup defined above wherein both $M^{(1)}$ and $M^{(2)}$ strictly satisfy the faithfulness axioms of Definition~\ref{def:faithful}, and assuming the slope of the insertion curve $C^{(2)}$ over the interval $[a_1, a_2]$ is bounded below by a strictly positive constant $\sigma > 0$, then:
\begin{equation}\label{eq:ins_lower}
\textnormal{AUC}_{\textnormal{Ins}}^{(2)} - \textnormal{AUC}_{\textnormal{Ins}}^{(1)}
\;\geq\; \tfrac{1}{2} \sigma \, (a_2 - a_1)^2.
\end{equation}
The accumulated gap remains strictly positive under any circumstance where method $2$'s supplementary positions contain individually informative predictive signals.
\end{proposition}

\begin{proof}
We begin by computing the explicit difference between the accumulated areas under the respective insertion curves. By definition, the Insertion AUC for method $k$ is the integral of its raw curve over the fractional domain $\alpha \in [0,1]$. Separating the baseline offset yields:
\[
\text{AUC}_{\text{Ins}}^{(k)} = \int_0^1 C^{(k)}(\alpha) \, d\alpha = C_{\text{base}} + \int_0^1 (C^{(k)}(\alpha) - C_{\text{base}}) \, d\alpha.
\]
Thus, the difference in AUC between the two methods reduces purely to the integral of the pointwise gap $\Delta(\alpha) = T(\alpha) - S(\alpha)$:
\[
\text{AUC}_{\text{Ins}}^{(2)} - \text{AUC}_{\text{Ins}}^{(1)} = \int_0^1 \Delta(\alpha) \, d\alpha.
\]
To isolate the region of divergence, we partition the domain of integration into three distinct intervals based on the active mask fractions $a_1$ and $a_2$:
\[
\int_0^1 \Delta(\alpha) \, d\alpha
= \underbrace{\int_0^{a_1} \Delta(\alpha) \, d\alpha}_{\text{Interval 1}}
+ \underbrace{\int_{a_1}^{a_2} \Delta(\alpha) \, d\alpha}_{\text{Interval 2}}
+ \underbrace{\int_{a_2}^1 \Delta(\alpha) \, d\alpha}_{\text{Interval 3}}.
\]
In Interval 1 ($[0, a_1]$), both methods sequentially reveal identical spatial-temporal features. Thus, $S(\alpha) = T(\alpha)$, rendering $\Delta(\alpha) = 0$, and ensuring the first integral evaluates exactly to zero. 
In Interval 3 ($[a_2, 1]$), the pointwise gap $\Delta(\alpha)$ is guaranteed to be non-negative because $T(\alpha) \geq S(\alpha)$ pointwise by the definition of the nested active sets. Therefore, the third integral contributes a value strictly greater than or equal to zero. 

Because the first integral is zero and the third is non-negative, we can establish a strict mathematical lower bound by analyzing Interval 2 exclusively:
\[
\text{AUC}_{\text{Ins}}^{(2)} - \text{AUC}_{\text{Ins}}^{(1)} \;\geq\; \int_{a_1}^{a_2} \Delta(\alpha) \, d\alpha.
\]
Within the fractional domain $(a_1, a_2]$, method $2$ continues to reveal genuinely informative positions, driving its curve upward. Conversely, method $1$ has entirely exhausted its active mask subset. By invoking Faithfulness Condition (iii) on $M^{(1)}$, we establish that any newly revealed positions past $a_1$ contribute exactly zero change to the model's confidence until $\alpha = 1$. Consequently, $S(\alpha)$ plateaus entirely: $S(\alpha) = S(a_1)$ for all $\alpha \in (a_1, 1)$. The gap function within Interval 2 thereby simplifies cleanly to $\Delta(\alpha) = T(\alpha) - S(a_1)$.

We are provided the hypothesis that the slope of $C^{(2)}$ (and equivalently $T$) over the interval $[a_1, a_2]$ is bounded below by the constant $\sigma > 0$. Because the curves intersect at the boundary such that $T(a_1) = S(a_1)$, we can project a linear lower bound onto the trajectory of $T(\alpha)$: $T(\alpha) - S(a_1) \geq \sigma(\alpha - a_1)$. 

Substituting this linear geometric bound back into the integral yields:
\[
\int_{a_1}^{a_2} \Delta(\alpha) \, d\alpha
\;\geq\; \int_{a_1}^{a_2} \sigma(\alpha - a_1) \, d\alpha.
\]
In other words, because the small mask flatlines and the big mask climbs at slope $\sigma$, they form a triangle of "free area" on the graph. The width of this triangle is $(a_2 - a_1)$. The height is $\sigma(a_2 - a_1)$. The area of a triangle is $\frac{1}{2} \cdot \text{base} \cdot \text{height}$. Evaluating this standard polynomial integral computes to $\tfrac{1}{2} \sigma (\alpha - a_1)^2$ evaluated from boundary $a_1$ to $a_2$, directly resulting in $\tfrac{1}{2} \sigma (a_2 - a_1)^2$. This derivation finalizes the formal quadratic bound presented in Eq.~\eqref{eq:ins_lower}. The established quadratic bound is structurally conservative because it entirely omits the accumulated area from the third interval $[a_2, 1]$.
\end{proof}

Consequently, the Insertion AUC metric exhibits an inherent structural bias that strictly favors spatially larger masks, even when controlling for per-position precision. Because a larger active mask possesses a wider fractional domain over which to incrementally increase the model's confidence, it mathematically accumulates a definitively greater area under the evaluation curve. In stark contrast, the Deletion AUC metric intrinsically penalizes an artificial inflation of mask size. Once a proposed mask successfully encompasses the entirety of the model's predictive evidence, further expansion of the active mask cannot mathematically depress the deletion curve any further. Consequently, Deletion AUC functions strictly as a measure of evidence coverage, remaining invariant to arbitrary increases in absolute mask size.

\begin{proposition}[Deletion AUC depends on coverage, not size]\label{prop:del_invariance}
Assume $M^{(1)}$ and $M^{(2)}$ represent faithful masks (per Definition~\ref{def:faithful}) that share the position importance ordering, subject to $0 < a_1 < a_2 < 1$. Suppose $M^{(1)}$ sufficiently covers the model's underlying evidence, formally defined such that there exists a fractional point $a^* \in (0, a_1]$ where the deletion curve $D^{(1)}$ perfectly converges to the baseline confidence: $D^{(1)}(a^*) = C_{\textnormal{base}}$. Under these conditions:
\begin{equation}\label{eq:del_invariance}
\textnormal{AUC}_{\textnormal{Del}}^{(2)} \;=\; \textnormal{AUC}_{\textnormal{Del}}^{(1)}.
\end{equation}
\end{proposition}

\begin{proof}
The deletion evaluation procedure at a given fractional threshold $\alpha$ dictates that the top-$\lfloor \alpha N \rfloor$ positions of the mask, ranked strictly in descending order, are removed and replaced with a baseline blur. The target-class probability is then recalculated. 

Because masks $M^{(1)}$ and $M^{(2)}$ perfectly share the underlying position importance ordering $\pi$, both methods inherently prioritize the exact same spatial-temporal features for initial deletion. Therefore, for any fraction $\alpha$ within the interval $[0, a_1]$, the deletion operations drop identical sets of pixels. This strictly guarantees that their respective deletion curves are pointwise equivalent throughout this initial domain: $D^{(1)}(\alpha) = D^{(2)}(\alpha)$ for all $\alpha \in [0, a_1]$.

Invoking the coverage hypothesis, the curve $D^{(1)}$ is known to reach the absolute baseline confidence $C_{\text{base}}$ at some fractional point $a^* \leq a_1$. By fundamental definition, a deletion curve is monotonically non-increasing and is strictly bounded from below by $C_{\text{base}}$. Because $D^{(1)}$ intersects its absolute minimum at $a^*$, it is geometrically constrained to remain perfectly horizontal thereafter: $D^{(1)}(\alpha) = C_{\text{base}}$ for all subsequent fractions $\alpha \in [a^*, 1)$.

Crucially, this exact geometric plateau must uniformly apply to $D^{(2)}$ over the interval $[a^*, 1)$. Because $D^{(1)}$ and $D^{(2)}$ mandate identical pixel removals up to $a_1$, and given that $a^* \leq a_1$, the curve $D^{(2)}$ has similarly been driven down to the baseline $C_{\text{base}}$ precisely at $\alpha = a^*$. Further obfuscation of features beyond $a^*$ cannot rationally reduce the model's confidence below the baseline probability. Thus, $D^{(2)}(\alpha) = C_{\text{base}}$ identically for all $\alpha \in [a^*, 1)$.

For the remaining fractional domain $\alpha \in [a_2, 1]$, both explanatory methods have completely exhausted their proposed active positions. They proceed to sequentially delete inactive (zero-weighted) positions governed entirely by the shared tie-breaking permutation. As theoretically established, both functions merely trace the $C_{\text{base}}$ minimum floor throughout this entire tail process until $\alpha = 1$.

Because $D^{(1)}(\alpha) = D^{(2)}(\alpha)$ over the domain $[0, a^*]$, and both curves strictly equate to $C_{\text{base}}$ over $[a^*, 1]$, we conclude that $D^{(1)} \equiv D^{(2)}$ globally across the entire closed interval $[0, 1]$. Integrating identical functional curves inevitably yields identical areas, cementing Eq.~\eqref{eq:del_invariance}.
\end{proof}

\paragraph{The asymmetry between insertion and deletion.} Once an explanatory mask achieves total coverage of the model's operative evidence, expanding its boundaries changes the deletion curve exclusively at thresholds $\alpha < a^*$, a domain where the curves of any structurally covering masks natively agree by construction. Deletion AUC thus rigorously evaluates whether a mask captures the prediction support, deliberately ignoring extraneous inclusions. The contrapositive statement proves equally diagnostic: if two masks produce substantially divergent deletion AUCs, we can mathematically conclude that at least one candidate fundamentally fails to capture the core evidence.

\par The structural advantage of Insertion AUC for larger masks fundamentally arises from the directional placement of the mask's ``tail'' (the large quantity of low-importance background pixels). We characterize this phenomenon as \emph{asymmetric early saturation}. Every explanation mask can be conceptually partitioned into a high-importance ``head'' (the core visual evidence) and a low-importance ``tail'' (the surrounding background context). The mathematical asymmetry between the two metrics lies entirely in how they integrate this tail.

\emph{The Insertion Phenomenon (Multiplying by High Confidence).} Under the insertion protocol, the mask reveals its head first. Once the critical evidence is revealed, the model's target-class confidence rapidly saturates near its maximum ($C \approx 1$). For a large, dense mask, the remaining evaluation steps consist of revealing its large tail. Because the confidence is already saturated, this tail is integrated over a high-confidence plateau. Geometrically, the tail creates a wide integration window that is multiplied by a high altitude, racking up strictly positive area simply because the mask takes longer to finish revealing pixels.

\emph{The Deletion Phenomenon (Multiplying by Zero).} Conversely, the deletion protocol hides the head first. Once the critical evidence is destroyed, the model is effectively blinded to the target class, and confidence crashes near its baseline minimum ($C \approx 0$). When the large mask proceeds to hide its massive tail, it operates within a ``dead zone.'' Integrating over this zero-confidence plateau contributes negligibly to the total area.  Ultimately, Deletion AUC strictly isolates the efficiency of the ``head'' because the tail is mathematically multiplied by zero once the evidence is destroyed. Insertion AUC, by building the evidence first, evaluates the tail at maximum confidence. This geometrically guarantees that larger masks receive artificially inflated scores purely as a function of their absolute pixel count, rather than the precision of their explanation.

Collectively, these two propositions rigorously harden the rhetorical claim presented in Section~\ref{sec:experiments}. Insertion AUC mathematically favors enlarged masks even when localized precision is controlled (Proposition~\ref{prop:ins_advantage}). Thus, achieving a superior insertion AUC remains highly ambiguous, consistent with either a highly precise compact mask or a loosely imprecise sprawling mask. Deletion AUC actively discriminates between these divergent scenarios (Proposition~\ref{prop:del_invariance}). Our proposed method's consistent dominance in deletion AUC across all three diverse datasets consequently serves as the far more resilient and diagnostic empirical finding.

Note, the presented propositions fundamentally presume \emph{exact} geometric faithfulness as defined in Definition~\ref{def:faithful}, an idealized property that no empirical explanation algorithm flawlessly achieves. The substantive analytical takeaway is qualitative rather than strictly numerical: Insertion AUC inherently harbors a documented geometric bias toward mask size, a vulnerability that Deletion AUC fundamentally resists. Real-world explanatory methods occupy a gradient between perfect faithfulness and arbitrary feature selection; however, the relative divergence between these two metrics on that continuum dependably tracks the limiting theoretical behavior established here.

\paragraph{Empirical check without the idealized faithfulness assumption.}\label{app:synthetic_masks}
Definition~\ref{def:faithful}(iii) is idealized. To test the size bias directly, we build synthetic masks in which size and precision vary independently: at precision $p$, a fraction $p$ of the active positions is taken from a gradient-saliency ranking and the rest are drawn uniformly at random. We grow each mask from $5\%$ to $60\%$ of all spatio-temporal positions at fixed precision.

\begin{table}[h]
\centering
\caption{Insertion vs.\ deletion AUC under a $12\times$ increase in mask size at fixed precision.}
\label{tab:synthetic_masks}
\small
\setlength{\tabcolsep}{4pt}
\begin{tabular}{@{}lcccccc@{}}
\toprule
Precision & Ins.\ (5\%) & Ins.\ (60\%) & Change & Del.\ (5\%) & Del.\ (60\%) & Change \\
\midrule
0.10 & 0.282 & 0.328 & $+16.4\%$ & 0.268 & 0.236 & $-11.8\%$ \\
0.30 & 0.293 & 0.396 & $+35.5\%$ & 0.258 & 0.228 & $-11.7\%$ \\
0.50 & 0.303 & 0.444 & $+46.6\%$ & 0.252 & 0.224 & $-11.1\%$ \\
0.70 & 0.311 & 0.484 & $+55.5\%$ & 0.244 & 0.223 & $-8.8\%$ \\
0.90 & 0.314 & 0.505 & $+60.8\%$ & 0.242 & 0.218 & $-9.9\%$ \\
\bottomrule
\end{tabular}
\end{table}

At every precision level, size moves insertion AUC $1.4$--$6.3\times$ more than deletion AUC, and the gap widens as precision increases, i.e., exactly for the compact, high-precision masks we aim to produce. This supports the qualitative conclusion of Propositions~\ref{prop:ins_advantage}--\ref{prop:del_invariance} without relying on Definition~\ref{def:faithful}(iii).

\subsection{Ablation of the noun/verb semantic split}\label{app:semantic_split}

We keep every architectural component and replace the separate noun and verb objectives with a single blended confidence objective, so neither mask receives an axis-specific semantic signal (50 examples from EGTEA data on the TSM architecture).

\begin{table}[h]
\centering
\caption{Effect of removing the semantic split. $\Delta$ = (w/o split) $-$ (full).}
\label{tab:semantic_split}
\small
\begin{tabular}{@{}lccc@{}}
\toprule
Metric & Full & w/o semantic split & $\Delta$ \\
\midrule
App.\ Del.\ AUC $\downarrow$ & 0.128 & 0.152 & $+0.024$ \\
Mot.\ Del.\ AUC $\downarrow$ & 0.234 & 0.285 & $+0.051$ \\
App.\ $\text{F}^{-}$ $\uparrow$   & 0.323 & 0.295 & $-0.028$ \\
Mot.\ $\text{F}^{-}$ $\uparrow$   & 0.656 & 0.616 & $-0.040$ \\
App.\ Ins.\ AUC $\uparrow$   & 0.630 & 0.673 & $+0.043$ \\
Mot.\ Ins.\ AUC $\uparrow$   & 0.703 & 0.778 & $+0.075$ \\
$\mm$ size (\%)              & $\approx 27$ & $36.3$ & $+9.3$ \\
\bottomrule
\end{tabular}
\end{table}

Removing the split worsens deletion AUC and $\text{F}^{-}$ on both axes. Insertion AUC rises, but together with a $\approx 34\%$ larger temporal mask, consistent with the size bias of insertion (App.~\ref{app:synthetic_masks}). Semantic supervision therefore contributes materially to the separation; the perturbation architecture alone does not determine it.

\subsection{Empirical cross-axis gradients}\label{app:leakage}

We measure, at convergence, the gradient magnitude of each loss with respect to each mask on 200 clips using a TSM model trained on the EGTEA dataset. The leak ratio is computed per clip as $\|\partial \mathcal{L}_{\text{verb}}/\partial \ma\| / \|\partial \mathcal{L}_{\text{noun}}/\partial \ma\|$ for $\ma$ and $\|\partial \mathcal{L}_{\text{noun}}/\partial \mm\| / \|\partial \mathcal{L}_{\text{verb}}/\partial \mm\|$ for $\mm$; $0$ would mean no cross-axis influence.

\begin{table}[h]
\centering
\caption{Gradient magnitudes at convergence. Ratios are computed per clip and then aggregated, so they are not the ratio of the column means.}
\label{tab:leakage}
\small
\begin{tabular}{@{}lccc@{}}
\toprule
Mask & Intended gradient & Cross-axis gradient & Leak ratio (mean / median) \\
\midrule
$\ma$ & $\|\partial\mathcal{L}_{\text{noun}}/\partial\ma\| = 1.45{\times}10^{-2}$ & $\|\partial\mathcal{L}_{\text{verb}}/\partial\ma\| = 1.38{\times}10^{-2}$ & 0.88 / 0.85 \\
$\mm$ & $\|\partial\mathcal{L}_{\text{verb}}/\partial\mm\| = 6.50{\times}10^{-1}$ & $\|\partial\mathcal{L}_{\text{noun}}/\partial\mm\| = 3.20{\times}10^{-1}$ & 5.31 / 1.23  \\
\bottomrule
\end{tabular}
\end{table}

Cross-axis gradients are of the same order as the intended ones. For $\mm$ the mean is inflated by clips whose intended verb gradient is nearly zero at convergence (median $1.07{\times}10^{-3}$), so the median is more representative. These measurements do not support a claim that gradients separate by semantic axis cleanly. Future work aims to improve the formulation with the use of stop-gradients to prevent gradient leakage.

\subsection{How much does optical flow drive the SSv2 motion mask?}\label{app:flow_ablation}

\begin{table}[h]
\centering
\caption{Flow-alignment ablation on 50 SSv2 clips (TSM). $r$: Pearson correlation between $\mm$ and per-frame flow magnitude.}
\label{tab:flow_ablation}
\small
\begin{tabular}{@{}lcc@{}}
\toprule
Metric & With flow ($\lambda_F{=}1.312$) & Without flow ($\lambda_F{=}0$) \\
\midrule
Mot.\ Ins.\ AUC $\uparrow$  & \textbf{0.334} & 0.186 \\
Mot.\ Del.\ AUC $\downarrow$ & 0.152 & \textbf{0.096} \\
Mot.\ $\text{F}^{+}$ $\uparrow$ & \textbf{0.465} & 0.175 \\
Mot.\ $\text{F}^{-}$ $\uparrow$ & 0.613 & \textbf{0.647} \\
$\mm$ size (\%) & 67.1 & 37.7 \\
$r(\mm, \text{flow})$ & 0.37 & -0.212 \\
\bottomrule
\end{tabular}
\end{table}

Flow alignment biases $\mm$ toward broader, more sufficient temporal explanations (higher insertion and $\text{F}^{+}$) at the cost of sparsity and deletion selectivity. The moderate correlation ($r=0.37$) shows that $\mm$ is influenced by, but does not simply copy, the flow prior. A model-derived replacement for the flow prior is left to future work.

\subsection{Seed variance and paired significance tests}\label{app:stats}

Each method is optimized with three seeds. For paired comparisons we compute, per clip, the difference between our method and each baseline, oriented so that positive values favor our method (lower is better for deletion, higher for insertion). We report the mean difference and the Wilcoxon signed-rank $p$-value.

\begin{table}[h]
\centering
\caption{Mean $\pm$ std over three seeds (TSM) on the SSV2 dataset.}
\label{tab:seeds}
\small
\setlength{\tabcolsep}{4pt}
\begin{tabular}{@{}llcccc@{}}
\toprule
Dataset & Method & App.\ Ins.\ $\uparrow$ & Mot.\ Ins.\ $\uparrow$ & App.\ Del.\ $\downarrow$ & Mot.\ Del.\ $\downarrow$ \\
\midrule
\multirow{6}{*}{SSv2}
& EP            & $0.321 \pm 0.002$ & - & $0.143 \pm 0.001$ & - \\
& EP + $\mm$    & $0.266 \pm 0.003$ & $0.402 \pm 0.007$ & $0.085 \pm 0.003$ & $0.399 \pm 0.007$ \\
& STEP          & $0.347 \pm 0.003$ & - & $0.172 \pm 0.001$ & - \\
& STEP + $\mm$  & $0.357 \pm 0.009$ & $0.488 \pm 0.001$ & $0.147 \pm 0.005$ & $0.444 \pm 0.012$ \\
& Unified Mask  & $0.409 \pm 0.008$ & - & $0.186 \pm 0.005$ & - \\
& Ours          & $0.318 \pm 0.007$ & $0.25 \pm 0.003$ & $0.11 \pm 0.004$ & $0.112 \pm 0.002$ \\
\bottomrule
\end{tabular}
\end{table}

\begin{table}[h]
\centering
\caption{Paired comparison of Ours against each baseline (TSM). Each cell gives the mean per-item difference, oriented so that positive values favor our method, and the Wilcoxon signed-rank $p$-value; $^{*}$: $p<0.05$. Items are clips for SSv2 ($n{=}200$) and EGTEA ($n{=}400$), and videos for EPIC-55 ($n{=}89$--$93$; per-segment scores averaged within each video and over seeds). Motion metrics are omitted for baselines without a motion mask, as in Table~\ref{tab:dataset_model_results}. Differences are computed on the seed-averaged per-item scores used for this test, so they can differ slightly from differences of the rounded means in Table~\ref{tab:dataset_model_results}.}
\label{tab:significance}
\resizebox{\textwidth}{!}{%
\begin{tabular}{@{}llcccc@{}}
\toprule
Dataset & Baseline & App.\ Ins.\ & Mot.\ Ins.\ & App.\ Del.\ & Mot.\ Del.\ \\
\midrule
\multirow{5}{*}{SSv2}
& EP + $\mm$   & \pdn{+0.052^{*}}{2.3{\times}10^{-3}} & \pdn{-0.146^{*}}{3.6{\times}10^{-9}} & \pdn{-0.028^{*}}{3.6{\times}10^{-5}} & \pdn{+0.286^{*}}{2.1{\times}10^{-28}} \\
& EP           & \pdn{-0.020}{0.30} & -- & \pdn{+0.002}{0.51} & -- \\
& STEP + $\mm$ & \pdn{-0.039^{*}}{8.6{\times}10^{-3}} & \pdn{-0.232^{*}}{6.7{\times}10^{-19}} & \pdn{+0.034^{*}}{0.033} & \pdn{+0.331^{*}}{8.0{\times}10^{-32}} \\
& STEP         & \pdn{-0.029^{*}}{0.021} & -- & \pdn{+0.059^{*}}{1.9{\times}10^{-4}} & -- \\
& Unified Mask & \pdn{-0.091^{*}}{4.4{\times}10^{-6}} & -- & \pdn{+0.072^{*}}{5.2{\times}10^{-15}} & -- \\
\midrule
\multirow{5}{*}{EGTEA}
& EP + $\mm$   & \pdn{+0.110^{*}}{3{\times}10^{-22}} & \pdn{+0.106^{*}}{7{\times}10^{-15}} & \pdn{+0.060^{*}}{6{\times}10^{-13}} & \pdn{+0.368^{*}}{3{\times}10^{-51}} \\
& EP           & \pdn{+0.089^{*}}{1{\times}10^{-16}} & -- & \pdn{+0.075^{*}}{7{\times}10^{-18}} & -- \\
& STEP + $\mm$ & \pdn{-0.010}{0.41} & \pdn{-0.028^{*}}{0.03} & \pdn{+0.115^{*}}{7{\times}10^{-43}} & \pdn{+0.449^{*}}{8{\times}10^{-59}} \\
& STEP         & \pdn{+0.006}{0.06} & -- & \pdn{+0.142^{*}}{3{\times}10^{-41}} & -- \\
& Unified Mask & \pdn{-0.075^{*}}{1{\times}10^{-5}} & -- & \pdn{+0.107^{*}}{2{\times}10^{-42}} & -- \\
\midrule
\multirow{5}{*}{EPIC-55}
& EP + $\mm$   & \pdn{+0.015^{*}}{0.034} & \pdn{+0.100^{*}}{8.4{\times}10^{-14}} & \pdn{+0.073^{*}}{2.5{\times}10^{-15}} & \pdn{+0.319^{*}}{1.4{\times}10^{-15}} \\
& EP           & \pdn{+0.029^{*}}{1.1{\times}10^{-4}} & -- & \pdn{+0.084^{*}}{1.1{\times}10^{-15}} & -- \\
& STEP + $\mm$ & \pdn{-0.152^{*}}{1.4{\times}10^{-14}} & \pdn{-0.061^{*}}{3.3{\times}10^{-10}} & \pdn{+0.122^{*}}{2.9{\times}10^{-16}} & \pdn{+0.427^{*}}{2.8{\times}10^{-16}} \\
& STEP         & \pdn{-0.138^{*}}{1.5{\times}10^{-12}} & -- & \pdn{+0.134^{*}}{8.1{\times}10^{-16}} & -- \\
& Unified Mask & \pdn{-0.151^{*}}{5.2{\times}10^{-16}} & -- & \pdn{+0.127^{*}}{1.5{\times}10^{-16}} & -- \\
\bottomrule
\end{tabular}}
\end{table}

\subsection{Input and mask resolution}\label{app:resolution}

Nothing in the framework fixes the input or mask resolution. At $512{\times}512$, peak memory grows by $\approx 4.1\times$ and runtime by $6.7$--$7.6\times$ relative to $224{\times}224$. Explanation quality at $512{\times}512$ cannot be assessed meaningfully with our classifiers as they were trained at $224{\times}224$ and their confidence on the \emph{unperturbed} clip drops from $0.855$ to $0.215$. Even in this setting, raising the seed-mask resolution from $32$ to $64$ improves motion insertion AUC from $0.307$ to $0.526$ and appearance insertion AUC from $0.142$ to $0.160$ on the EGTEA dataset with the TSM model, suggesting that finer seed masks help localize small objects in egocentric video when paired with a classifier trained at the target resolution.

\subsection{Auditing example on EGTEA Gaze+}\label{app:audit}

We group 200 unfiltered EGTEA test clips by predicted verb and rank classes by few active $\mm$ frames and high verb confidence. A nominally temporal class whose motion explanation repeatedly collapses to almost no frames while confidence stays high is a candidate for a static-appearance shortcut.

\begin{table}[h]
\centering
\caption{Verb classes ranked by few active motion frames (of 8) and high confidence (TSM). }
\label{tab:audit}
\small
\begin{tabular}{@{}lccc@{}}
\toprule
Predicted verb & Clips & Mean active $\mm$ frames & Mean verb confidence \\
\midrule
Wash        & 5  & 0.80 & 0.895 \\
Cut         & 10 & 1.40 & 0.849 \\
Move around & 4  & 1.50 & 0.776 \\
Close       & 3  & 1.67 & 0.875 \\
\bottomrule
\end{tabular}
\end{table}

For \emph{wash}, the motion mask keeps fewer than one frame on average while the classifier is $\approx 90\%$ confident, suggesting the verb is recognized largely from static cues (e.g., sink, running water) rather than the washing motion. The mask alone does not establish a shortcut. It produces a testable hypothesis: freezing all frames to the single most appearance-salient frame should leave \emph{wash} confidence largely intact, whereas a motion-dependent class should drop. Conversely, preserving appearance while altering the trajectory tests whether the model depends on motion at all.
\subsection{Hyperparameters}\label{app:hyperparams}
\begin{figure}
    \centering
    \includegraphics[width=0.7\linewidth]{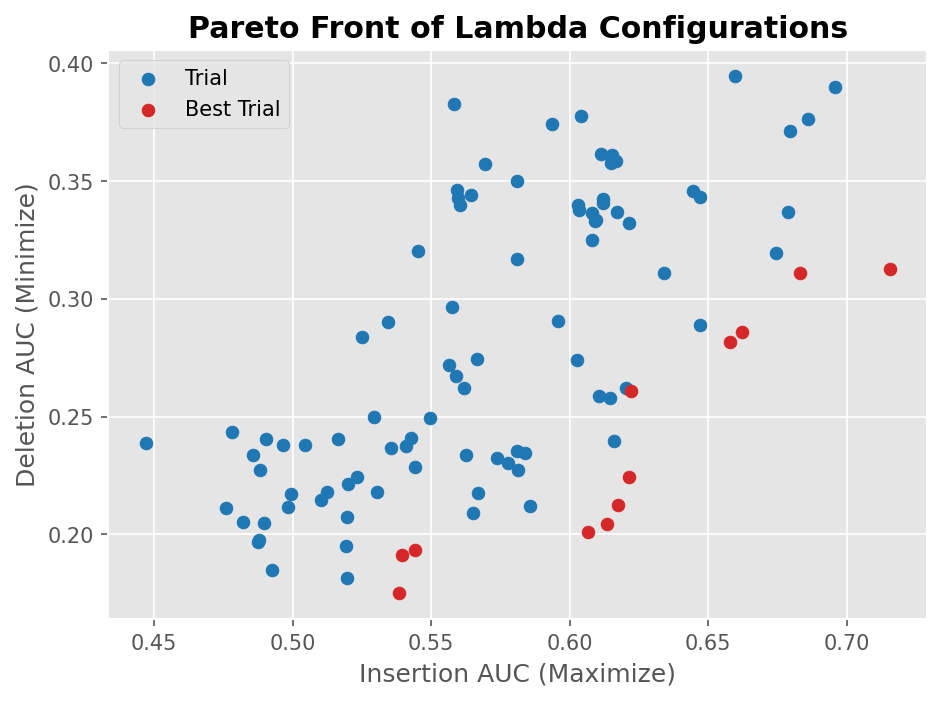}
    \caption{Figure showing how hyperparameters $(\lambda_A, \lambda_M, \lambda_S, \lambda_D)$ were picked with a held out validation set using a combination of grid search and Bayesian optimization on Optuna for minimizing deletion AUC and maximizing Insertion AUC across 100 trials on the EGTEA Gaze+ dataset with the TSM model.}
    \label{fig:pareto}
\end{figure}
Figure \ref{fig:pareto} shows how  hyperparameters $(\lambda_A, \lambda_M, \lambda_S, \lambda_D)$ were picked for the EPIC-Kitchens 55 dataset using a held out validation set, across 100 trials. Table \ref{tab:hyperparameters} shows the hyperparameters that resulted from the Bayesian optimization on a held-out set.
\begin{table}[h!]
\centering
\small
\setlength{\tabcolsep}{4pt}
\caption{Optimization hyperparameters for our disentangled mask
  $(M_a, M_m)$. Loss weights $(\lambda_A, \lambda_M, \lambda_S,
  \lambda_D, \lambda_F)$ were tuned on a held-out split via
  Bayesian optimization. Optimizer settings,
  schedule, and temperature anneal are identical across datasets and
  backbones.}
\label{tab:hyperparameters}
\begin{tabular}{lccc}
\toprule
\multicolumn{4}{l}{\textbf{(a) Loss weights and target areas (per dataset)}} \\
\midrule
                                       & EPIC-Kitchens-55 & EGTEA Gaze+ & Something-Something V2 \\
\midrule
$\lambda_A$ \ \ (M$_a$ area)           & $13.85$ & $0.3841$ & $0.133$ \\
$\lambda_M$ \ \ (M$_m$ area)     & $9.44$ & $0.786$ & $0.263$ \\        
$\lambda_S$ \ \ (L$_K$ smoothness)     & $0.143$ & $0.0107$ & $0.0877$ \\
$\lambda_D$ \ \ (disentanglement)      & $7.026$ & $38.88$ & $38.88$ \\
$\lambda_F$ \ \ (flow alignment) & N/A    & N/A     & $1.312$ \\        
\addlinespace
$\nu_a$ \ \ (target spatial area for M$_a$)      & $0.20$   & $0.53$   & $0.6$ \\
$\nu_m$ \ \ (target temporal area for M$_m$)     & $0.50$   & $0.61$   & $0.6$ \\
\midrule
\multicolumn{4}{l}{\textbf{(b) Optimizer and schedule (shared across datasets)}} \\
\midrule
\multicolumn{2}{l}{Optimizer}                     & \multicolumn{2}{l}{Adam, $(\beta_1, \beta_2)=(0.9, 0.999)$, $\epsilon{=}10^{-8}$} \\
\multicolumn{2}{l}{Trainable parameters}          & \multicolumn{2}{l}{$\boldsymbol{\ell}_a \in \mathbb{R}^{T \times 32 \times 32}$,\ \ $\boldsymbol{\ell}_m \in \mathbb{R}^{T}$ (logits)} \\
\multicolumn{2}{l}{Learning rate}                 & \multicolumn{2}{l}{$\alpha = 0.05$ (constant)} \\
\multicolumn{2}{l}{Weight decay}                  & \multicolumn{2}{l}{$0$} \\
\multicolumn{2}{l}{Iterations}                    & \multicolumn{2}{l}{$N{=}500$ per clip} \\
\multicolumn{2}{l}{Logit initialization}          & \multicolumn{2}{l}{$\boldsymbol{\ell}_a, \boldsymbol{\ell}_m \leftarrow -1.0$\ \ ($\sigma(-1){\approx}0.27$)} \\
\multicolumn{2}{l}{Mask upsample}                 & \multicolumn{2}{l}{$32{\times}32 \rightarrow 224{\times}224$ via Gaussian (factor $7\times$)} \\
\multicolumn{2}{l}{$L_K$ kernel}                  & \multicolumn{2}{l}{spatial $31{\times}31$ (stride $31$), temporal $3$ (stride $1$)} \\
\midrule
\multicolumn{4}{l}{\textbf{(c) Temperature annealing schedule}} \\
\midrule
\multicolumn{2}{l}{Mask parameterization}         & \multicolumn{2}{l}{$M = \sigma(\tau_i \cdot \boldsymbol{\ell})$} \\
\multicolumn{2}{l}{Annealing form}                & \multicolumn{2}{l}{linear in iteration} \\
\multicolumn{2}{l}{$\tau_i$}                      & \multicolumn{2}{l}{$\tau_i = \tau_{\min} + (\tau_{\max} - \tau_{\min}) \cdot \frac{i}{N - 1}$} \\
\multicolumn{2}{l}{$\tau_{\min} \rightarrow \tau_{\max}$} & \multicolumn{2}{l}{$1.0 \rightarrow 10.0$ over $500$ iterations} \\
\multicolumn{2}{l}{Final-mask temperature}        & \multicolumn{2}{l}{$\tau_{\max} = 10.0$ (binarizing)} \\
\bottomrule
\end{tabular}
\end{table}

\subsection{Hyperparameter sensitivity}\label{app:sensitivity}

We vary one hyperparameter at a time around our tuned configuration, holding all others fixed, on 20 clips per dataset with three optimization seeds per setting (Table~\ref{tab:sensitivity}). Seed-to-seed standard deviations are small relative to the trends (at most $0.035$ on EGTEA and $0.062$ on SSv2), so the effects below are not noise.

\paragraph{Area budgets.} The appearance budget $v_a$ has almost no effect: across a $3\times$ range all four AUCs vary by at most $0.006$, consistent with the upper-bound area loss being inactive when the optimized mask is already smaller than the budget. The motion budget $v_m$ leaves insertion unchanged but trades away deletion selectivity as it grows (App.\ Del.\ $0.162 \rightarrow 0.224$), and motion Fidelity$^-$ drops at the largest budget ($0.71 \rightarrow 0.59$).

\paragraph{Separation weight $\lambda_D$.} $\lambda_D$ is the most influential parameter and controls a trade-off between sufficiency and selectivity on both datasets: increasing it raises insertion AUC (SSv2 Mot.\ Ins.\ $0.042 \rightarrow 0.431$) and motion Fidelity$^+$ ($0.04 \rightarrow 0.64$), while deletion AUC also increases. Very small $\lambda_D$ yields near-zero deletion AUC on SSv2, but only because the motion mask collapses: motion insertion and Fidelity$^+$ fall to almost zero. A low deletion AUC on its own is therefore not evidence of a good explanation, and our tuned value lies in the regime where both sufficiency and selectivity are reasonable.

\paragraph{Flow weight $\lambda_F$.} On SSv2, across a $50\times$ range of $\lambda_F$ App.\ Ins.\ stays within $0.25$--$0.33$ and Mot.\ Ins.\ within $0.30$--$0.42$; only the weakest flow prior ($\lambda_F{=}0.1$) noticeably lowers motion insertion, and values between $1$ and $2$ form a plateau. The flow term is used only for single-head classifiers, so it has no effect on EGTEA (identical results at every value).

\paragraph{Optimization steps.} Insertion AUC increases monotonically with the number of steps, with diminishing returns beyond our setting of $500$ on SSv2 (App.\ Ins.\ $+0.011$ from $500$ to $750$). Deletion AUC rises with more steps as the masks become more sufficient, and $100$ steps are clearly too few (SSv2 appearance Fidelity$^+$ $0.55$ vs.\ $0.97$ at $500$).

\begin{table}[h]
\centering
\caption{Hyperparameter sensitivity (TSM; 20 clips $\times$ 3 seeds, mean AUC). One hyperparameter is varied at a time; $^\dagger$ marks the tuned value used for all other rows. This study uses its own clip subset, so absolute values are not directly comparable to Table~\ref{tab:dataset_model_results}.}
\label{tab:sensitivity}
\small
\setlength{\tabcolsep}{5pt}
\begin{tabular}{@{}llrcccc@{}}
\toprule
Dataset & Param. & Value & App.\ Ins.\ $\uparrow$ & Mot.\ Ins.\ $\uparrow$ & App.\ Del.\ $\downarrow$ & Mot.\ Del.\ $\downarrow$ \\
\midrule
\multirow{20}{*}{EGTEA}
& \multirow{5}{*}{$v_a$}
  & 0.26 & .699 & .730 & .193 & .270 \\
& & 0.40 & .696 & .729 & .196 & .267 \\
& & 0.53$^\dagger$ & .700 & .730 & .192 & .270 \\
& & 0.66 & .694 & .729 & .197 & .270 \\
& & 0.79 & .696 & .731 & .198 & .270 \\
\cmidrule(l){2-7}
& \multirow{5}{*}{$v_m$}
  & 0.31 & .695 & .727 & .162 & .267 \\
& & 0.46 & .693 & .727 & .179 & .270 \\
& & 0.61$^\dagger$ & .700 & .730 & .192 & .270 \\
& & 0.76 & .698 & .732 & .212 & .277 \\
& & 0.92 & .700 & .734 & .224 & .283 \\
\cmidrule(l){2-7}
& \multirow{5}{*}{$\lambda_D$}
  & 19.4 & .604 & .686 & .122 & .244 \\
& & 29.2 & .658 & .710 & .159 & .256 \\
& & 38.9$^\dagger$ & .700 & .730 & .192 & .270 \\
& & 48.6 & .710 & .771 & .214 & .284 \\
& & 58.3 & .724 & .787 & .231 & .295 \\
\cmidrule(l){2-7}
& \multirow{5}{*}{Steps}
  & 250 & .639 & .649 & .153 & .212 \\
& & 375 & .682 & .706 & .177 & .238 \\
& & 500$^\dagger$ & .700 & .730 & .192 & .270 \\
& & 625 & .706 & .750 & .196 & .292 \\
& & 750 & .731 & .769 & .207 & .309 \\
\midrule

\multirow{14}{*}{SSv2}
& \multirow{5}{*}{$\lambda_D$}
  & 1 & .105 & .042 & .018 & .026 \\
& & 5 & .128 & .042 & .027 & .032 \\
& & 20 & .265 & .246 & .088 & .128 \\
& & 38.9$^\dagger$ & .281 & .395 & .134 & .209 \\
& & 50 & .297 & .431 & .133 & .230 \\
\cmidrule(l){2-7}
& \multirow{5}{*}{$\lambda_F$}
  & 0.1 & .327 & .297 & .077 & .133 \\
& & 1 & .284 & .397 & .134 & .193 \\
& & 1.31$^\dagger$ & .281 & .395 & .134 & .209 \\
& & 2 & .305 & .421 & .139 & .222 \\
& & 5 & .253 & .373 & .133 & .168 \\
\cmidrule(l){2-7}
& \multirow{4}{*}{Steps}
  & 100 & .115 & .016 & .031 & .010 \\
& & 250 & .226 & .195 & .107 & .090 \\
& & 500$^\dagger$ & .281 & .395 & .134 & .209 \\
& & 750 & .292 & .448 & .143 & .237 \\
\bottomrule
\end{tabular}
\end{table}

\subsection{Compute and timing}\label{app:compute}

All optimization-based explanation methods were evaluated using a single NVIDIA DGX. Because our framework and the baselines act as post-hoc explainers applied to a frozen classifier, no model retraining is required. The computational cost per example scales primarily with the number of optimization iterations ($N_{\text{iter}} = 500$) and the complexity of the forward and backward passes through the chosen video backbone.

On the EGTEA Gaze+ and EPIC-Kitchens-55 datasets using the TSM backbone ($T=8$ frames), our full factored method requires an average of 11.7 seconds per video clip. The computational overhead of our nearest-active-frame temporal replacement strategy adds only minor latency compared to standard single-mask baselines: STEP requires 11.4s (12.8s with the $M_m$ temporal extension), and Extremal Perturbation requires 10.3s (10.4s with temporal extension). The Unified Mask ablation, which optimizes only a single spatiotemporal mask without the appearance-motion factorization, bypasses the temporal gating entirely and is consequently the fastest at 7.3s per example.

For Something-Something~V2, which inherently demands more complex spatial-temporal processing, our full method requires 18.1 seconds per example. Given the evaluation subset sizes ($N=400$ for EGTEA, $N=9385$ for EPIC-Kitchens-55, and $N=200$ for SSv2), generating the complete suite of explanations for a single method on a single dataset takes roughly 1 to 2 hours for EGTEA and ~32hours for EPIC-Kitchens. Factoring in all baseline evaluations, multiple backbones (TSM and TSFormer), and hyperparameter grid searches, the total compute budget for the experiments reported in this paper was approximately 400 to 500 GPU hours.

\end{document}